\documentclass[lettersize,journal]{IEEEtran}
 
\usepackage{times}
 
\usepackage{multicol}
 
\usepackage{colortbl}
\usepackage{hhline}
\usepackage{multirow}

\usepackage[bookmarks=true]{hyperref}
 
\usepackage[utf8]{inputenc}
\usepackage{helvet}

\usepackage{caption}
\usepackage{subcaption}
\usepackage{rotating}
 
\usepackage{tablefootnote,multirow}
\usepackage{diagbox}
\usepackage{float}
\usepackage{dblfloatfix}
 
\usepackage{amsmath,amsfonts,mathtools,siunitx, amsthm}
\usepackage{amssymb}
\usepackage{graphicx}
\usepackage{adjustbox}
\usepackage{epstopdf}
\usepackage{xcolor}
\usepackage{bm}
\usepackage{lipsum}  
\usepackage[url=false, sorting=none]{biblatex}
\newcommand{\R}{\mathbb{R}}
\newcommand{\nn}{\nonumber}
\newcommand{\ddi}[1]{\frac{\partial #1}{\partial i}}

\usepackage{float}
\usepackage{dblfloatfix}
\usepackage{accents} 
\newcommand{\Oh}{\mathcal{O}}
\newcommand{\sqb}[1]{\left[ {#1}\right] }
\newcommand{\parb}[1]{\left( {#1}\right) }

\newtheorem{proposition}{Proposition}

\newtheorem{lemma}{Lemma}

\newtheorem{remark}{Remark}

\usepackage{algpseudocode}
\usepackage{algorithm}
\newcommand{\norm}[1]{\left\lVert{#1}\right\rVert}

\IEEEoverridecommandlockouts
 
\usepackage[left=0.75in,bottom=0.78in,right=0.751in,top=0.755in]{geometry}
 
\begin{document}
\title{\vspace{0.1in} Learning High Dimensional Demonstrations Using Laplacian Eigenmaps}
 
\author{Sthithpragya Gupta, Aradhana Nayak, Aude Billard
\thanks{The authors are with Learning Algorithms and Systems Laboratory (LASA), EPFL, 1015 Lausanne, Switzerland (sthithapragya.gupta@epfl.ch, aradhana.nayak@epfl.ch, aude.billard@epfl.ch)}}

\maketitle
 
\begin{abstract}

This article proposes a novel methodology to learn a stable robot control law driven by dynamical systems. The methodology requires a single demonstration and can deduce a stable dynamics in arbitrary high dimensions. The method relies on the idea that there exists a latent space in which the nonlinear dynamics appears quasi linear. The original nonlinear dynamics is mapped into a stable linear DS, by leveraging on the properties of graph embeddings. We show that the eigendecomposition of the Graph Laplacian results in linear embeddings in two dimensions and quasi-linear in higher dimensions. The nonlinear terms vanish, exponentially as the number of datapoints increase, and for large density of points, the embedding appears linear. We show that this new embedding enables to model highly nonlinear dynamics in high dimension and overcomes alternative techniques in both precision of reconstruction and number of parameters required for the embedding. We demonstrate its applicability to control real robot tasked to perform complex free motion in space.
\end{abstract}
 
\IEEEpeerreviewmaketitle
 
\section{Introduction}
\noindent Robots are often required to execute novel tasks that involve convoluted motions in high dimensions. The traditional methods of robot programming to achieve this may require varying levels of user input that may be infeasible and impractical. In these settings, learning from demonstration (LfD) offers an elegant solution to facilitate robot programming and learning novel trajectories for task completion \cite{ravichandar_recent_2020-1}. The goal is to replace the manual process of programming the robot with an automatic programming process whose input consists of demonstrations performed by an expert. The challenges this paper sets out to address are 1) to learn a stable control law from a single expert demonstration as it is difficult and expensive to generate a large input data set and 2) to stabilize highly nonlinear dynamics in high dimensions.
 
We set forth that the underlying control law is driven by a dynamical system (DS) and that the demonstration is an instance of one path integral, albeit possibly noisy one, of this DS. If the motion has a target, the DS consists of a vector field with asymptotic stability convergence properties to a single attractor, the target of the motion. If the vector field is properly identified from the single path integral, the DS guarantees generalization of the dynamics to unseen areas of the state space.  
 

 \begin{figure}[H]
    \centering
    \begin{subfigure}{0.45\textwidth}
    \centering
    \includegraphics[width=\linewidth, trim={16cm 0 15cm 0},clip]{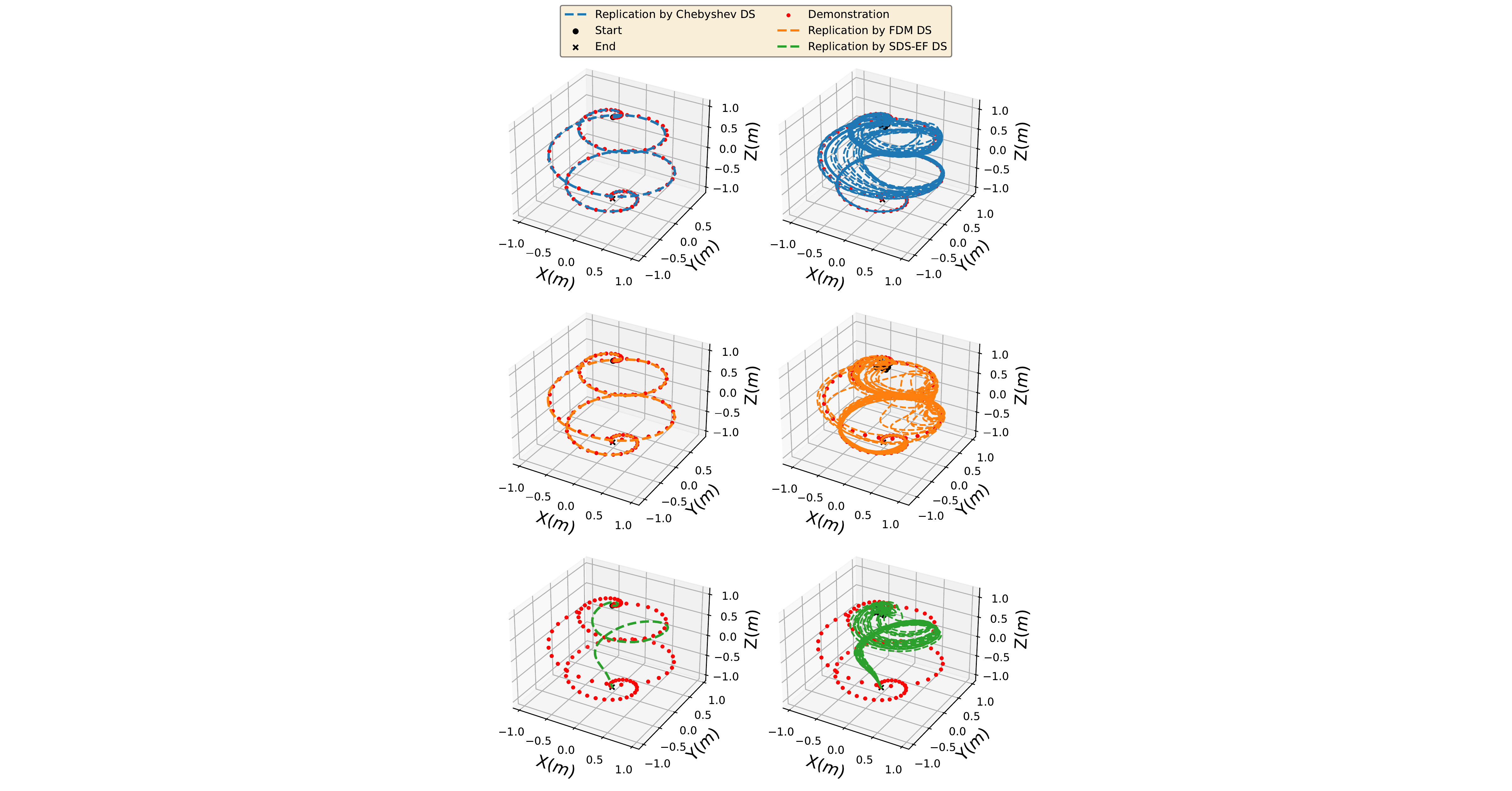}
    \caption{\scriptsize Superimposed path integrals generated from initializing the DS from the original starting point (left) and from 20 random initialisations (right)}
    \label{fig:replications}
    \end{subfigure}
    \centering
    \begin{subfigure}{0.5\textwidth}
    \centering
    \includegraphics[width=0.5\textwidth]{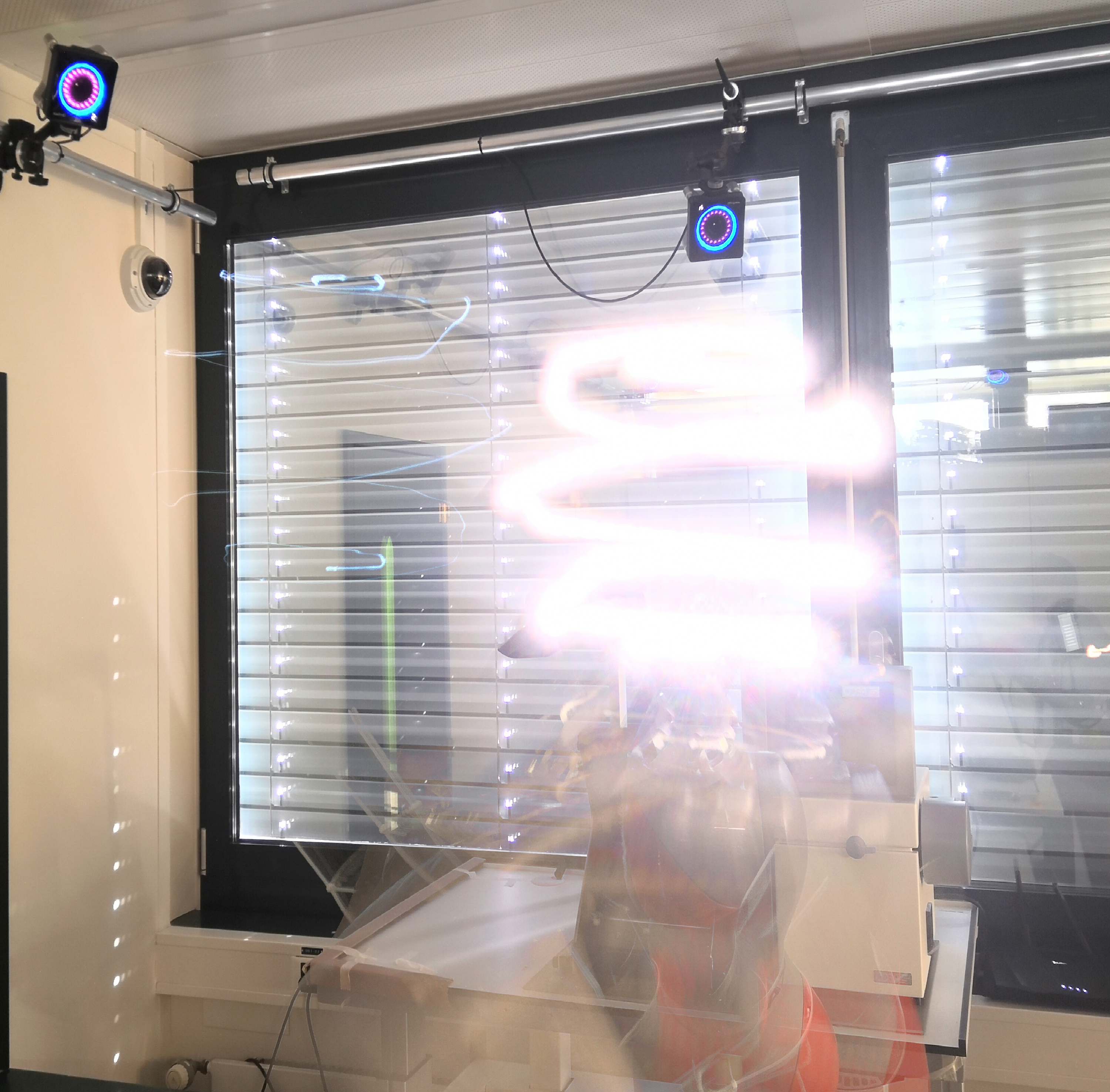}
    \caption{\scriptsize Light painting of the replication generated by the DS learnt using our method}
    \end{subfigure}
    \label{fig:comparison_intro}
    \caption{\scriptsize (Top) Comparative reconstruction of a 3D spiralling dynamics, when using the proposed Chebyshev latent space embedding for learning the dynamics to two state of the art embeddings through diffeormorphism (FDM \cite{perrin_fast_2016} and Euclideanising flows (SDS-EF) in \cite{rana_euclideanizing_2020},   using an unstable 3D spiral demonstration of complexity $c=7$. (Bottom) Example of one trajectory generated by the learned dynamical systems and reproduced on a real robot.).}
    \label{fig1}
\end{figure}

This paper proposes a novel methodology to learn the vector field of a stable DS by transforming the original $n$-dimensional task space to a `latent space’ where the dynamics appear linear or quasi linear. The rationale is this latent linear embedding would make it easier to stabilise the DS in original space. To obtain this embedding, we express the DS through a graph representation of points in the demonstration and compute an eigendecomposition of the associated graph Laplacian. We show that a subset of the eigenvectors of the Graph Laplacian describe a space in which the dynamics is linear if 2D and quasi linear for dimension higher than 2. We further show that the non-linearity of the embedding vanish with $O(1/N^2)$, $N$ being the number of data points. The dynamics in this embedding follow a Chebyshev polynomial, and hence, we refer to our new embedding as {\em Chebyshev DS} in the rest of this paper. 

To reconstruct the dynamics in original space, we learn a diffeomorphism between the latent space and the demonstration space, using the  fast diffeomorphic matching (FDM) algorithm proposed in \cite{perrin_fast_2016}. When comparing our embedding to the use of other latent spaces, we obtain better reconstruction of highly nonlinear motions and better generalization of the dynamics when initialized away from the initial demonstration, see example in 
Figure \ref{fig1}.
 
The paper is organised as follows. In Section \ref{background}, we review closely related work and introduce the problem formulation in Section \ref{section:problem_def}. Section \ref{section:graph_formulation} revisits Belkin and Niyogi's formulation of Laplacian eigenmaps and introduce its application on our use case. The main theoretical results related to the construction of latent space embedding are documented in Section \ref{section:chebFormulation}. Section \ref{section:learning} describes the application of the diffeomorphism learning approach to our latent space embedding and choice of hyperparameters. Section \ref{section:results} reports on the quantitative and qualitative comparative evaluations in simulation and robot implementation experiments, that evaluate the efficacy of the proposed latent space. We end the paper by presenting our conclusions in Section \ref{section:conclusions}.
 
\section{Background \& Related Work}\label{background}
\begin{figure}[ht]
    \centering
    \includegraphics[width=0.9\linewidth, trim={11cm 0cm 11cm 0cm},clip]{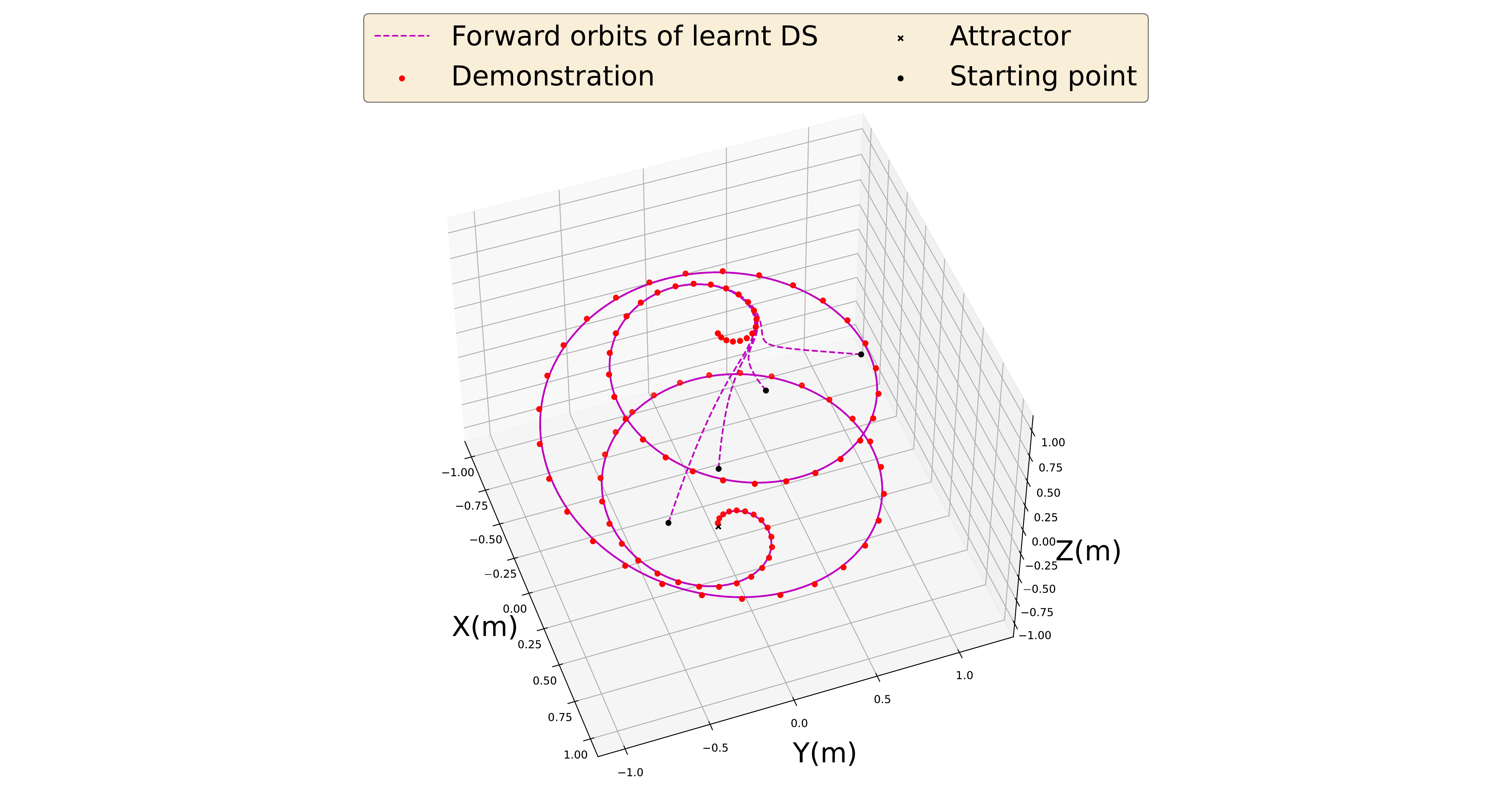}
    \caption{Forward orbits generated by the DS learnt using DMP}
    \label{fig:DMP_performance}
\end{figure}

\noindent DS-based LfD was first tackled with dynamical motion primitives (DMPs) in \cite{schaal_is_1999}. In the classical DMP approach, temporal disturbances are handled by a phase variable, and a demonstration in the joint space is represented by a DMP per joint. Therefore, it is difficult to accurately synchronise the phase across several DMPs. \cite{rai2014learning} show that even when the DS is stable, the vector field governing the flow of the DS intersects itself as shown in Fig. \ref{fig:DMP_performance}. Furthermore, it is not robust to spatial disturbances during roll-out, as shown in \cite{ravichandar2017learning}.
 
Basis functions are employed to model the dynamics that encapsulate the nonlinear features of the vector field governing the DS. The basis functions are usually Gaussians (\cite{khansari-zadeh_learning_2011}, \cite{mohammad_khansari-zadeh_learning_2014}, \cite{ravichandar2017learning}) of appropriate dimension that are parameterised so as to preserve the stability of the DS in accordance to a chosen or optimal Lyapunov function. In the stable estimator of dynamical systems (SEDS) approach (\cite{khansari-zadeh_learning_2011-1}), the Lyapunov function is fixed as the square of the distance to the attractor. In \cite{mohammad_khansari-zadeh_learning_2014}, the underlying Lyapunov function is modelled by choosing it from a set of weight sum of asymmetric quadratic functions. In \cite{neumann_learning_2015}, $\tau-$SEDS, the demonstration space is transformed by a diffeomorphism so that SEDS can be applied. \cite{figueroa2018physically} propose a linear parameter varying DS (LPV-DS) to improve the performance of SEDS using parameterised quadratic Lyapunov functions. \cite{chaandar2019learning} and \cite{ravichandar2017learning} propose the existence of an underlying positive definite contraction metric, and the Gaussian mixture model (GMM) is selected under constraints imposed by the contraction metric. However, only a certain class of contraction metrics is considered. 
 
The limitations of the above approaches are more pronounced as the complexity of the demonstration increases. This is due to an inaccurate estimation of the underlying Lyapunov function that yields a poor replication of the nonlinearity in the demonstration. This can be overcome either by learning the Lyapunov function together with the dynamics with a neural network as in \cite{NEURIPS2019_0a4bbced} or by learning a diffeomorphism between a latent space and demonstration space. The characteristic feature of the latent space is that the transformed DS in the latent space is either linear or highly simplified. There are two distinct approaches to learning the aforementioned diffeomorphism. The first is a geometric approach that learns from just one demonstration either with FDM \cite{perrin_fast_2016} or by the large deformation diffeomorphism metric mapping approach in \cite{joshi2000landmark}. The latter is hard to invert to the original space, and the trajectories in the latent space are generated by a linear DS of the form $\dot{x}=-x$ in \cite{perrin_fast_2016}. While it is true that a diffeomorphism transforms the original DS into a simplified latent space DS, the explicit formulation of the latent space DS is unknown.
 
In the second approach, this diffeomorphism is learnt from multiple demonstrations such as stable dynamical system learning using Euclideanising flows (SDS-EF) in \cite{rana_euclideanizing_2020} and its modification to stochastic systems in \cite{urain2020imitationflow}. In SDS-EF, the diffeomorphism is expressed using function approximators based on single-layer neural networks, wherein the layer resembles a Gaussian kernel. The formulation relies on the fact that the DS in the latent space is the negative gradient descent of a chosen potential function. As a generative model is learnt, the number of demonstrations required to train the neural network is very high compared to the single diffeomorphism learning approach in \cite{perrin_fast_2016}. This is why SDS-EF performs poorly when a single demonstration is provided.
 
In summary, both approaches for diffeomorphism-based reconstruction of stable dynamics rely on an underlying DS in the latent space. In this work, we propose to improve the algorithm in \cite{perrin_fast_2016} by choosing coordinates in the latent space based on a graph representation of the demonstration. \begin{figure*}[ht]
    \centering
    \includegraphics[width=\linewidth]{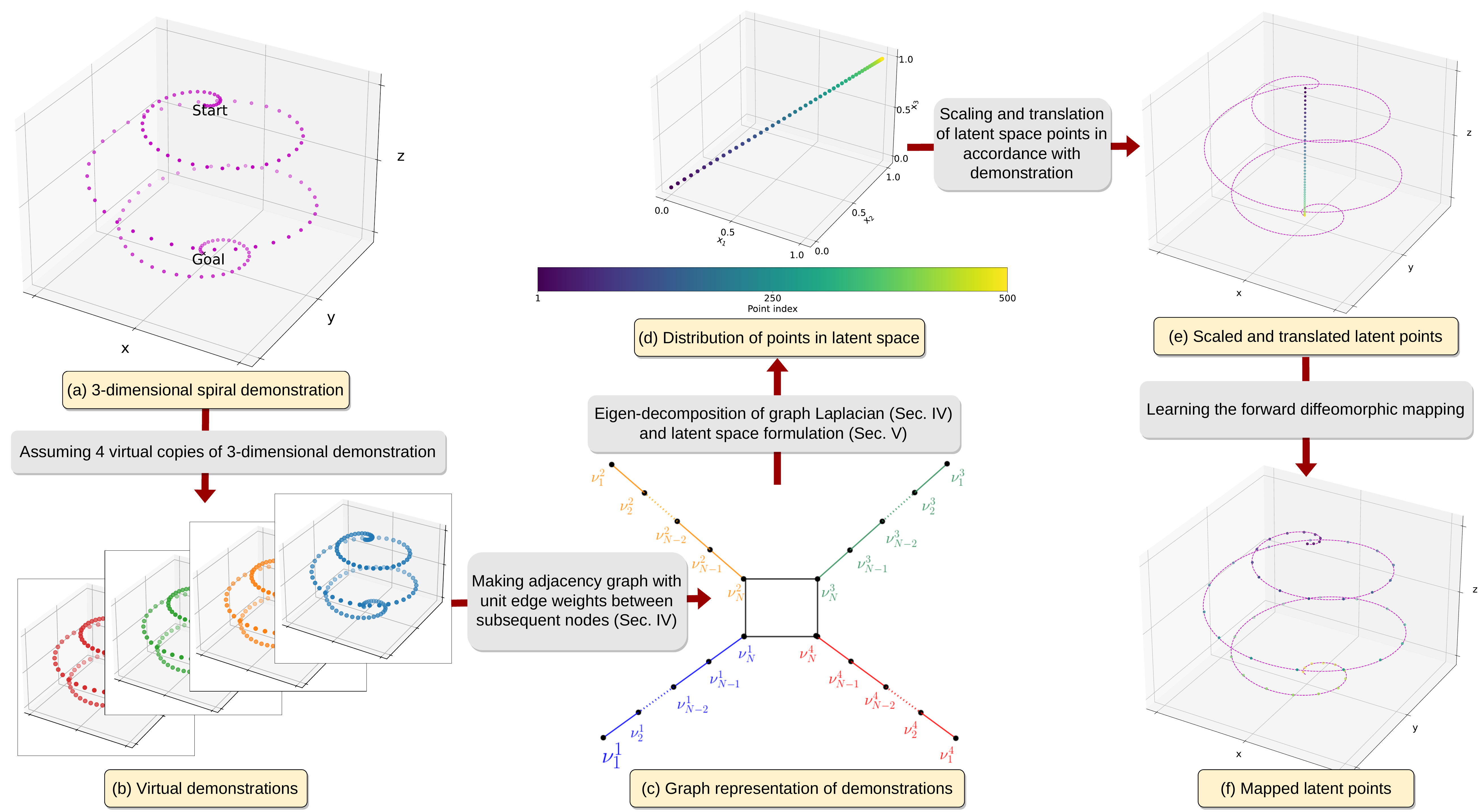}
    \caption{Framework of the procedure used to prepare the latent space and learn the diffeomorphic map. (a) An unstable 3D spiral demonstration of complexity $c=7$ (Section \ref{section:results}); (b) Assuming four copies of the 3-dimensional demonstration (Section \ref{section:graph_formulation}); (c) Graph representation of demonstrations (Section \ref{section:graph_formulation}); (d) Distribution of points in latent space corresponding to a single demonstration (Section \ref{eq:chebyshevFormulation}); (e) Latent space points scaled and translated between the start and end points of the demonstration; (f) Mapped points after learning the forward mapping (Section \ref{section:learning})}
    \label{fig:framework}
\end{figure*}
 
The approach of using graph Laplacian for dimensionality reduction was introduced in \cite{belkin2003laplacian}. In our formulation of latent space, we construct an unweighted graph whose nodes represent points in the demonstration and whose edges connect the neighbouring points. The coordinates in latent space correspond to the entries of certain eigenvectors of the Laplacian. We observe that the dynamical system in latent space is quasi-linear by analysing the chosen eigenvector entries. This simplifies the latent space dynamics.
 
In \cite{FicheraBillard2022}, the problem has been addressed for a demonstration evolving in a $2D$ space. Therefore, the method is only applicable to learn a demonstration in joint space for a robot with two joints or to learn a demonstration in a $2$-dimensional task space. In most common robotic applications, such as pick and place objects, the task space is higher than two dimensions. This paper extends the proposed algorithm in \cite{FicheraBillard2022} to learn complex demonstrations in high dimensions given a single demonstration for training. 
 
The main contributions of this work:
\begin{enumerate}
    \item Learning a stable, highly nonlinear autonomous dynamical system (DS) from a single demonstration.
    \item Identifying a latent space wherein the dynamics are quasi-linear, easing learning of highly nonlinear dynamics. 
    \item Proving that the DS in latent space becomes linear for a large number of datapoints $N$, as the nonlinear terms decrease with $\Oh(1/N^2)$. 
    \item Demonstrating that the proposed latent space embedding requires fewer parameters than other approaches with similar accuracy at reconstruction and accounts better for high nonlinearities. 
    \end{enumerate}
 
\section{Problem Setup}\label{section:problem_def}
\noindent The provided demonstration consists of $N$ position--velocity pairs ${\{(\bm{y}^i,\dot{\bm{y}}^i)\}}_{i=1}^N$ wherein the positions are $\bm{y}^i \in \R^n$ and the velocities are $\dot{\bm{y}}^i \in \R^n$. These position--velocity pairs are sampled at a constant frequency from a trajectory that is the path integral of a DS. The DS is defined as follows:
\begin{enumerate}
    \item The nonlinear function $f: \R^n \to \R^n$ describing the DS maps the $n$-dimensional position vector $\bm{y}^i$ to the corresponding velocity vector $\dot{\bm{y}}^i$ by the ordinary differential equation (ODE)
\begin{align}
    \dot{y} = f(y), \quad y \in \R^n \label{DS_actual}
\end{align}
\item The DS is globally asymptotically stable at $y^\star \in \R^n$, and, therefore, $ \lim_{t\to \infty} y(t)= y^*$ for all path integrals $y(t)$ of the DS.
\end{enumerate}
Consider a diffeomorphism $\psi: \R^n \to \R^n$ that transforms the demonstration space into a latent space. The coordinates in the latent space are defined as $x \coloneqq \psi(y)$. The time-parameterised path integrals $y(t)$ of the DS in \eqref{DS_actual} transformed by $\psi$ are denoted as $x(t)$ and defined as $x(t) \coloneqq \psi(y(t))$. These path integrals are solutions to the following ODE describing the DS in the latent space:
\begin{align}
    \dot{x}=  g(x) =: \frac{\partial \psi}{\partial y}\bigg\rvert_{ \psi^{-1}(x)}  f(\psi^{-1}(x)) \label{DS_latent}
\end{align}
The DS in \eqref{DS_latent} is also globally and asymptotically stable at $x^\star \coloneqq \psi(y^\star)$ as
\[  x^\star = \psi \big[\lim_{t\to \infty} y(t)\big]= \lim_{t\to \infty} \psi(y(t))= \lim_{t\to \infty} x(t)
\]
Therefore, one can recover the DS in \eqref{DS_actual}, given the DS in the latent space in \eqref{DS_latent} by the following transformation:
\begin{align}
    \dot{y} = \frac{\partial \psi^{-1}}{\partial x}\bigg\rvert_{\psi(y)}g(\psi(y)) \label{transform_DS}
\end{align}
Hence, the DS of latent space in \eqref{DS_latent} and the diffeomorphism $\psi$ are sufficient to reconstruct the dynamics of the DS in \eqref{DS_actual}. The main focus of this work is to construct an accurate and stable latent space DS. Then, state-of-the-art methods are applied to learn $\psi$, and the DS in the original space is recovered by the transformation in \eqref{transform_DS}.
 
In the following section, we proceed by obtaining a \emph{stable} latent space DS from the observed points $(\bm{y^i},\dot{\bm{y}}^i)$. In addition to being stable, we also show that the DS is \emph{quasi-linear}. The coordinates of the latent space are given by $n$ carefully chosen eigenvectors of a graph Laplacian. In the next section, we formulate the graph from the observed points following the approach in \cite{belkin2003laplacian}. We make specific assumptions on the structure of the graph that ensure desired properties of stability and quasi-linearity in the DS in the latent space.
 
\section{Generating a Graph from a Single Demonstration} \label{section:graph_formulation}
 
\noindent Consider $K$ copies of the single given demonstration. The graph is denoted by $G$ and has $N \times K$ nodes. A node of the graph is denoted as $\nu_i^j$, $i\in\{1,\dotsc, N\}$, $j\in \{1,\dotsc K\}$ and corresponds to the $i$th position--velocity pair of the demonstration in the $j$th copy of the demonstration set. The edges are given as node pairs $\{(\nu_i^k, \nu_{i+1}^k )\}_{i=1}^{N}$ and $k=1,\dotsc, K$. The edge weight is unity. The nodes labeled as 
\begin{align*}
   & \{\nu_1^1,\dotsc\nu_{N}^1,\nu_1^2,\dotsc,\nu_{N}^2,\dotsc, \nu_1^K,\dotsc,\nu_{N}^K \} \sim\\ &\{1,2, \dotsc {N}\times K\}
\end{align*}
The K nodes in the set $\lbrace \nu^1_{p_1},\dots, \nu^K_{p_K} \rbrace$ form a \emph{cyclic graph} (or simple circuit). Fig. \ref{fig:framework} (c) is the graph representation of a demonstration set with $N$ observations and $K=4$ copies of the demonstration. The $N$th observation of each copy is represented by the node $\nu_N^k$ and corresponds to the last point of the demonstration. This last point is deemed to be at, or close to, the attractor $y^\star$ of the DS in \eqref{DS_actual}. By construction, $G$ preserves the local connectivity across the points conveyed in the demonstration. The edge weights for $G$ are formulated as:
 
 
\begin{align}
     &\text{edge}\{\nu_i^k, \nu_{j}^l\} = \begin{cases}
     1 \:\text{if}\: k = l \:\text{\&}\: \|i-j\| = 1 \:\text{or}\: i=j=N\\
     0 \:\text{otherwise}
    \end{cases}
\end{align}
 The eigenvalues and eigenvectors are defined as solutions to the generalised eigenvector problem,
\begin{align}
L(G) \bm{u}^l = \lambda \bm{u}^l, \quad l=1,\dotsc, N\times K, \quad \bm{u}^l \in \R^{N\times K}\label{u_eqn}
\end{align}
where the graph Laplacian is denoted by $L(G)$ and preserves the neighbourhood information of the demonstration set. It has been used extensively in manifold learning for extracting latent representations of data in high dimensions and in spectral clustering. 
 
In this section, we study the entries of eigenvectors of $L(G)$ and show the existence of $n$ eigenvectors that form the basis of the latent space. In this latent space, the embedded dynamics are both stable and quasi-linear. In Subsection \ref{subsection:graph_laplacian_preparation}, the eigenvector entries are explicitly computed from the eigenequation \eqref{u_eqn} and are shown to be polynomial functions of the eigenvalues. In Subsection \ref{subsection:LS_from_Laplacian}, we show the existence of at least $n$ eigenvalues in the spectrum of $L(G)$, which, if distinct, differ from each other by a term in $\Oh(1/N^2)$. We conclude the section by showing that the entries of eigenvectors corresponding to these $n$ eigenvalues are quasi-linear.

\vspace{-3mm} 
\subsection{Analysis of Laplacian} \label{subsection:graph_laplacian_preparation}
 
We make a few observations on the structure of $L(G)$ to simplify the study of its eigenvectors. By definition, $L(G) = D(G) - A(G)$, where the matrices $D(G)$ and $A(G)$ denote the degree and adjacency matrices of $G$. $L(G)$ can be expressed in terms of the block circulant matrix $J$ as follows:
\begin{align*}
&L(G)=2I_{{N}\times K} -J,\\
&J = bcirc(B_1, B_2, \underbrace{ zeros({N}), \dotsc, zeros({N})}_\text{(K-3) times}, B_2)\\
&B_1 =\begin{pmatrix}
          1 & 1 & 0 & \cdots & 0 &0 \\
          1 & 0 & 1 & \cdots & 0 &0 \\
          \vdots & \vdots &\cdots & \ddots & \vdots &\vdots\\
          0 & 0 & 0 & \cdots & 0 & 1\\
          0 & 0 & 0 & \cdots & 1 & -1
        \end{pmatrix} \in \R^{{N} \times {N}},\\  
&B_2 = \begin{pmatrix}
                                     zeros({N}-1) & \mathbf{0} \\
                                     \mathbf{0} & 1
                                   \end{pmatrix}\in \R^{{N} \times {N}} 
\end{align*}

where $bcirc$ denotes a block circular matrix defined as
\[bcric(B_1,B_2,\dotsc,B_L)\coloneqq \begin{pmatrix} B_1 & B_2 &\dotsc & B_L\\
B_L & B_1 &\dotsc& B_{L-1}\\
\vdots & \vdots &\ddots &\vdots\\
B_2& B_{L}&\dotsc  & B_1
\end{pmatrix},
\]
for $ L\geq 3$, and $ zeros({N})$ denoting an ${N}\times {N}$ matrix of zeros, $\mathbf{0}$ denoting an ${N}$ vector of zeros.

The block circulant structure inside the $L(G)$ allows us to identify many eigenvalues in the spectrum of $L(G)$ that repeat with algebraic multiplicity equal to $2$. The following proposition gives the exact number of such repeating eigenvalues.
 
\begin{proposition}\label{propn2}
Denote the number of eigenvalues of $L(G)$ having an algebraic multiplicity of $2$ by $\Lambda$. We have
\begin{equation}
    \Lambda = \begin{cases}
    \frac{K}{2}-1 &  \text{if }K=2j, \quad j=\{1,2,\dotsc\} \\
    \frac{K-1}{2} & \text{if }K=2j-1, \quad j=\{2,3,\dotsc\}
    \end{cases}
\end{equation}
 
\end{proposition}
 
\begin{proof}
From Section 3.1 of \cite{tee2007eigenvectors}, we observe that the eigenvalues and corresponding eigenvectors of $J$ are determined by the following $K$ equations, each giving $N$ eigenvalues and vectors.
\begin{align}
    H_j v &= \lambda_j v, \quad j \in \{0,\dotsc, K-1 \} \nonumber \\
    H_j &:= B_1+ B_2 (\rho_j+ \rho_j^{K-1})\label{Hj_eqn}
\end{align} 
where $\rho_j$ is the $n$-th root of $1$ given by
\begin{align*}
\rho_j = \exp^{\frac{i 2\pi j}{K}},  j \in \{0,\dotsc, K-1 \}
\end{align*}
Observe that $\rho_j^{(K-1)} =\rho_j^{-1} $ for all $j$. Further,
\begin{align*}
\rho_j+\rho_j^{-1} &= 2 \cos\left( \frac{2 \pi j}{K} \right) = 2 \cos\left( \frac{2 \pi (K-j)}{K} \right) \\
&= \rho_{K-j}+\rho_{K-j}^{-1}, \quad j \in \{1,\dotsc, K-1 \}
\end{align*}
Because eigenvalues and vectors of $L(G)$ and $J$ are identical, the conclusion follows from \eqref{Hj_eqn}.
\end{proof}
 
 
Each eigenvector $\bm{u}^l$ of $L(G)$ possesses a peculiar structure, as observed by the following lemma.
 
\begin{lemma}\label{eigvec_repeat}
Denote the first $N$ entries of the eigenvector $\bm{u}^l$ by $u^l \in \R^N$. Then we have
\begin{align}
    \bm{u}^l = \begin{pmatrix} u^l &
    \rho u^l & \rho^2 u^l & \cdots &
    \rho^{K-1} u^l\end{pmatrix} ^\top \label{eigvec_struc}
\end{align}
where $\rho$ is the $K$th root of unity.
\end{lemma}
\begin{proof}
The result follows from the fact that eigenvectors of $L(G)$ and $J$ are identical. The eigenvectors of the block circulant matrix $J$ with $K$ blocks, each of size $N$, are given by \eqref{eigvec_struc} following the result from Section 3.1 in \cite{tee2007eigenvectors}.
\end{proof}
 
For simplicity in notation, we henceforth denote $u^l$ and $\bm{u}^l$ as $u$ and $\bm{u}$, respectively. The following proposition shows that the first $N$ entries of $\bm{u}$ can be expressed as polynomial functions of their corresponding eigenvalues.

\begin{proposition}\label{propn3}
The entries $u_i$ of $u= [u_1, \dotsc, u_N]$ corresponding to the eigenvalue $\lambda$ are expressed as combination of Chebychev polynomials $T_i$ and $V_i$ as
\begin{align}
    & u_{i}(\lambda) = u_{1}\left[T_i(1-\frac{\lambda}{2}) -  \frac{\lambda}{2}V_{i-1}\:(1-\frac{\lambda}{2})\right]
\label{eq:chebyshevFormulation}
\end{align}
$T_i$ and $V_i$ are Chebychev polynomials of first and second kind defined as
\begin{align}
    &T_i(\theta) = \operatorname{cos}(i \operatorname{arccos}(\theta)) \label{ti}\\
     &V_i(\theta) = \frac{\operatorname{sin}((i+1) \operatorname{arccos}(\theta))}{\operatorname{sin}(\operatorname{arccos}(\theta))}\label{vi}
\end{align}
 
\end{proposition}
 
\begin{proof}
From the eigenequation \eqref{u_eqn}, we have
\begin{align}\label{latentFormulation}
     &u_2 =(1-\lambda)u_1\\\nn
     &u_i = (2-\lambda)u_{i-1}- u_{i-2}, \quad i =3,\dotsc,N-1 
\end{align}
The entries $u_i$ of any vector of $L(G)$ can be modelled as a combination of Chebyshev polynomials of the first and second kind that are defined as $T(\theta)$ and $V(\theta)$, respectively, as follows:
\begin{align}
\begin{split}
    & T_0(\theta) = 1\\
    & T_1(\theta) = \theta\\
    & T_{i}(\theta) = 2\theta T_{i-1}(\theta) - T_{i-2}(\theta)
\end{split}
    \label{eq:firstChebyshev}
\end{align}
which yields $T_i(\theta)$ in \eqref{ti}. Similarly, 
\begin{align}
\begin{split}
    & V_0(\theta) = 1\\
    & V_1(\theta) = 2\theta\\
    & V_{i}(\theta) = 2\theta V_{i-1}(\theta) - V_{i-2}(\theta)
\end{split}
    \label{eq:secondChebyshev}
\end{align}
which yields $V_i(\theta)$ in \eqref{vi}. Choosing $\theta=(1-\lambda/2)$ in both \eqref{eq:firstChebyshev} and \eqref{eq:secondChebyshev}, we obtain the recursion in \eqref{latentFormulation} with the combination proposed in \eqref{eq:chebyshevFormulation}.
\end{proof}

\subsection{Latent space arising from Laplacian eigenmaps}\label{subsection:LS_from_Laplacian}
The set of basis vectors of the latent space is the set of first $N$ components of eigenvectors corresponding to the $n$ smallest non-zero repeating eigenvalues of $L(G)$ for $G$ with $K=n+1$ paths. These particular eigenvector components are chosen as they have monotonically increasing or decreasing entries, as shown in the following lemma. We shall soon see that this property also proves the stability of the DS in the latent space. We shall also see the need for choosing $K=n+1$.
 
\begin{lemma}\label{Lemma2}
Consider the eigenvector $\bm{u}$ corresponding to the eigenvalue $\lambda$ s.t.
\begin{equation}\label{mon_cond_lambda}
    0 < \lambda \leq 2\left[ 1 - \text{cos}\left(\frac{\pi}{N-\frac{1}{2}} \right) \right]
\end{equation}
The entries $u_i$ of $u= [u_1, \dotsc, u_N]$, which are the first $N$ components of the eigenvector $\bm{u}$, either increase or decrease monotonically s.t.
\[  u_j \ge (\le) u_{j+1}, \quad j =\{1,\dots, N\}
\]
\end{lemma}
 
\begin{proof}
In Appendix \ref{app3}
\end{proof}
 
Because the sequence $\{ u_j\}_{j=1}^N$ is bounded, by the monotone convergence theorem, the sequence converges to the lower bound if $ u_j \ge  u_{j+1}$ and to the upper bound if $ u_j \le u_{j+1}$. This shows that the DS in the latent space resulting from the first $N$ entries of $\bm{u}$ is stable. 
 
Lemma \ref{Lemma2} holds for eigenvectors corresponding to eigenvalues satisfying a certain upper bound. In the following lemma, we show that such eigenvalues always exist in the spectrum of $L(G)$.
 
\begin{lemma}\label{Lemma3}
The smallest repeating $K-1$ eigenvalues of $L(G)$, denoted $\lambda_j$, $j=1...K-1$, have as upper bound:
  \begin{equation}\label{small_eig_L_bds}
      \lambda_j(L) < 2\left( 1- \cos \left(\frac{\pi}{N}\right) \right), \quad j = \{1,\dotsc,(K-1) \} 
  \end{equation} 
\end{lemma}
\begin{proof}
In Appendix \ref{app4}
\end{proof}
 
From Lemma \ref{Lemma3}, in order for us to have $n$ eigenvalues satisfying \eqref{small_eig_L_bds}, we choose the number of paths $K=n+1$ in $G$. The first $N$ components of the eigenvectors corresponding to these $n$ eigenvalues are monotonic by Lemma \ref{Lemma2}. Choosing these vectors as basis vectors of the latent space, we obtain a \emph{stable} representation of the demonstration in the latent space.
 
Next we tackle the issue of \emph{quasi linearity} of the embedded demonstration in the latent space. Observe that the upper bound in \eqref{small_eig_L_bds} belongs to $\Oh(1/N^2)$ by considering a power series expansion of the cosine function in \eqref{small_eig_L_bds}: 
  \begin{equation}\label{small_eig_L}
    \lambda_j(L) < 2\sqb{\parb{\frac{\pi}{N}}^2 \frac{1}{2!}- \parb{\frac{\pi}{N}}^4\frac{1}{4!}+ \dotsc } \in \Oh(1/N^2) 
  \end{equation}
with $ j = \{1,\dotsc, n\}$ for a graph with $K=n+1$. Therefore, the upper bound holds for at least $n$ eigenvalues. In the following proposition, we show that the eigenvectors corresponding to these eigenvalues also differ entry-wise by a small term that belongs to $\Oh(1/N^2)$. These eigenvectors form the basis of the $n$-dimensional latent space.
 
 
 
\begin{proposition}\label{propn5}
Consider the graph $G$ constructed with $K=n+1$ copies of the demonstration. The vector space given by the first $N$ components of $n$ eigenvectors corresponding to the smallest $n$ repeating eigenvalues of $L(G)$ is the latent space. In this space, the graph embedding of the demonstration is linear up to an order $\Oh(1/N^2)$. In particular, $\norm{u^l-u^m}_2 \in \Oh(1/N^2)$ for $u^l$ and $u^m$ (in \eqref{eigvec_repeat}) corresponding to $\bm{u^l}$ and $\bm{u^m}$ (in \eqref{u_eqn}) respectively for repeating eigenvalues of $L(G)$.
\end{proposition}
\begin{proof}
In Appendix \ref{app5}
\end{proof}
\begin{remark}
Observe from \eqref{eq:chebyshevFormulation} that the eigenvectors corresponding to eigenvalues repeating with algebraic multiplicity equal to $2$ differ only in the scaling factor $u_0$. Given a $2D$ demonstration, the latent space is composed of the first $N$ components of the eigenvector corresponding to the smallest repeating eigenvalue. Therefore, we obtain a linear embedding of the dynamics in the latent space. This result appears in \cite{FicheraBillard2022}. 
\end{remark}
 
Note that the construction of the graph (and hence the latent space) does not require the position--velocity values from the supplied demonstration. Because $G$ assumes a binary structure, only $N$ (length of the demonstration) and $n$ (dimensions of demonstration space) are required.
 
\section{Formulation of Latent Space} \label{section:chebFormulation}
 
\noindent We denote the latent representation of our data set in the embedding as $X\coloneqq \{\bm{x}^1, \bm{x}^2 \hdots \bm{x}^N\}$ and drop the scaling term $u_0$ in \ref{eq:chebyshevFormulation} for simplicity. Each $\bm{x}^i$ is $n$-dimensional.
 
\begin{align}
    \begin{split}
        & \bm{x}^i = (x_{i,1}, x_{i,2}, \dots, x_{i,n})^T, \quad i =\{ 1,\dotsc,N\}\\ 
        & x_{i,j} = T_i(1-\frac{\lambda_j}{2}) -  \frac{\lambda_j}{2}V_{i-1}\:(1-\frac{\lambda_j}{2}), \quad j =\{ 1,\dotsc,n\}
    \end{split}
    \label{eq:chebyshevInputInit}
\end{align}
with $T_i$ and $V_i$ being defined in Equations \eqref{ti} and \eqref{vi}, respectively. From Proposition \ref{propn5}, we know that $\bm{x}^i$ are quasi co-linear, and the interpoint spacing between successive points increases monotonically. 
As previously outlined, the latent space is prepared by considering $K$ copies of the demonstration. The formulation from \ref{eq:chebyshevInputInit} can further be simplified as:
\begin{align}
    \begin{split}
    x_{i,j} &= \cos(i \arccos( 1 - \frac{\lambda_j}{2})) - \frac{\lambda_j}{2} \frac{\sin(i\arccos(1-\frac{\lambda_j}{2}))}{\sin(\arccos(1 - \frac{\lambda_j}{2}))}\\
    &= \operatorname{cos}(i a_j) - b \operatorname{sin}(i a_j)\nn =  \sqrt{b_j^2 + 1}\operatorname{sin}(\gamma - i a_j) \label{eq:inputDiscrete}
    \end{split}
\end{align}
where $a_j= \operatorname{arccos}(1-\frac{\lambda_j}{2})$, $b_j= \frac{\lambda_j}{2 sin(a_j)}$ and $\gamma_j = \operatorname{arcsin}(\frac{1}{\sqrt{b_j^2+1}})$
It follows from the above formulation that
\begin{align}
& \bm{x^1} = \bm{1_n^T}-\bm{\lambda}\\
& \bm{x^N} = \sqrt{\bm{b}^2 + 1}\operatorname{sin}(\bm{\gamma} - N \bm{a})\label{eq:latentSpaceSingleDim}
\end{align}
where $\bm{\lambda} = (\lambda_1, \lambda_2 \dots \lambda_n)^T$, $\bm{a} = (a_1, a_2 \dots a_n)^T$, $\bm{b} = (b_1, b_2 \dots b_n)^T$ and $\bm{\gamma} = (\gamma_1, \gamma_2 \dots \gamma_n)^T$. To model the input for learning the diffeomorphism, the order of the points in $X$ is reversed. With a slight abuse of notation, we refer to this reversed set also as $X \in \mathbb{R}^{N \times n}$. $X$ hereby represents a latent space embedding of the system that starts at $\bm{x}_0$ (formerly $\bm{x}^N$) and terminates at $\bm{x}^*$ (formerly $\bm{x}^1$), as shown in Figure \ref{fig:framework} (c).

\section{Learning the Diffeomorphism}\label{section:learning}
\noindent We propose to follow the FDM approach in \cite{perrin_fast_2016} to learn the diffeomorphism $\psi:\R^n \to \R^n$ between the latent space prepared in Section \ref{section:chebFormulation} and the demonstration space such that $\psi(\bm{x}^i) = \bm{y}^i$ for every $i \in \{1,\dotsc,N\}$. We collectively denote points in the latent space by $X\coloneqq \{\bm{x}^i\}_{i=1}^{N}$ and points in the demonstration space by $Y\coloneqq \{\bm{y}^{i}\}_{i=1}^{N}$. The diffeomorphism is expressed as a composition of $M$ individual diffeomorphisms as $\psi = \psi_M\circ\dotsc\circ \psi_1$, wherein each $\psi_j:\R^n\to \R^n$ is a diffeomorphism in the layer $j$ of the algorithm.
\subsection{Hyperparameter tuning}
The hyperparameters associated with learning the diffeomorphism in accordance with the FDM approach are $\mu$, $\beta$ and $M$: 
\begin{itemize}
    \item $M$: Number of diffeomorphisms constituting the forward map $\psi$. A lower $M$ value is favourable as it lowers the chance of overfitting and the computational complexity associated with inverting the learnt forward map
    \item $\mu \in (0,1)$: Constrains each diffeomorphism to remain invertible
    \item $\beta \in (0,1]$: Behaves like the learning rate while learning the forward map. A lower value is favourable for stable convergence and reduced chance of overfitting
\end{itemize}
The metric used to measure the relevance of the hyperparameter combinations ($\mu, \beta$, $M$) was the mean-squared error (MSE) between the demonstration $Y$ and the final deformed input space $\psi(X)$. The combination with the least number of successive diffeomorphisms ($M$) was selected from amongst the hyper-parameter combinations ($\mu, \beta$, $M$), producing a normalised MSE less than $1e-5$ (unless having slightly more diffeomorphisms significantly reduced MSE). Amongst the combinations with the same number of successive diffeomorphisms, there are two possible candidates -- the combination the with least $\beta$ value and the combination with the lowest MSE. If none of the combinations could produce an MSE lower than $1e-5$, the combination with the lowest MSE was selected. 
 
The hyperparameters for learning the diffeomorphism associated with Euclideanising flows are $L$ and $n_f$:
\begin{itemize}
    \item $L$: The number of layers constituting the network. A lower $L$ value is favourable as it lowers the chance of overfitting.
    \item $n_f$: The number of features per layer. Once again, a lower value is favourable to a lower chance of overfitting.
\end{itemize}
The demonstration data were split into train and test sets, with the training set constituting $70\%$ of the data. The metric used to measure the relevance of the hyperparameter combinations ($L, n_f$) was the MSE on the test set.
 
\begin{figure}[ht]
    \centering
    \includegraphics[width=\linewidth,trim={6cm 0 7.5cm 1cm}, clip]{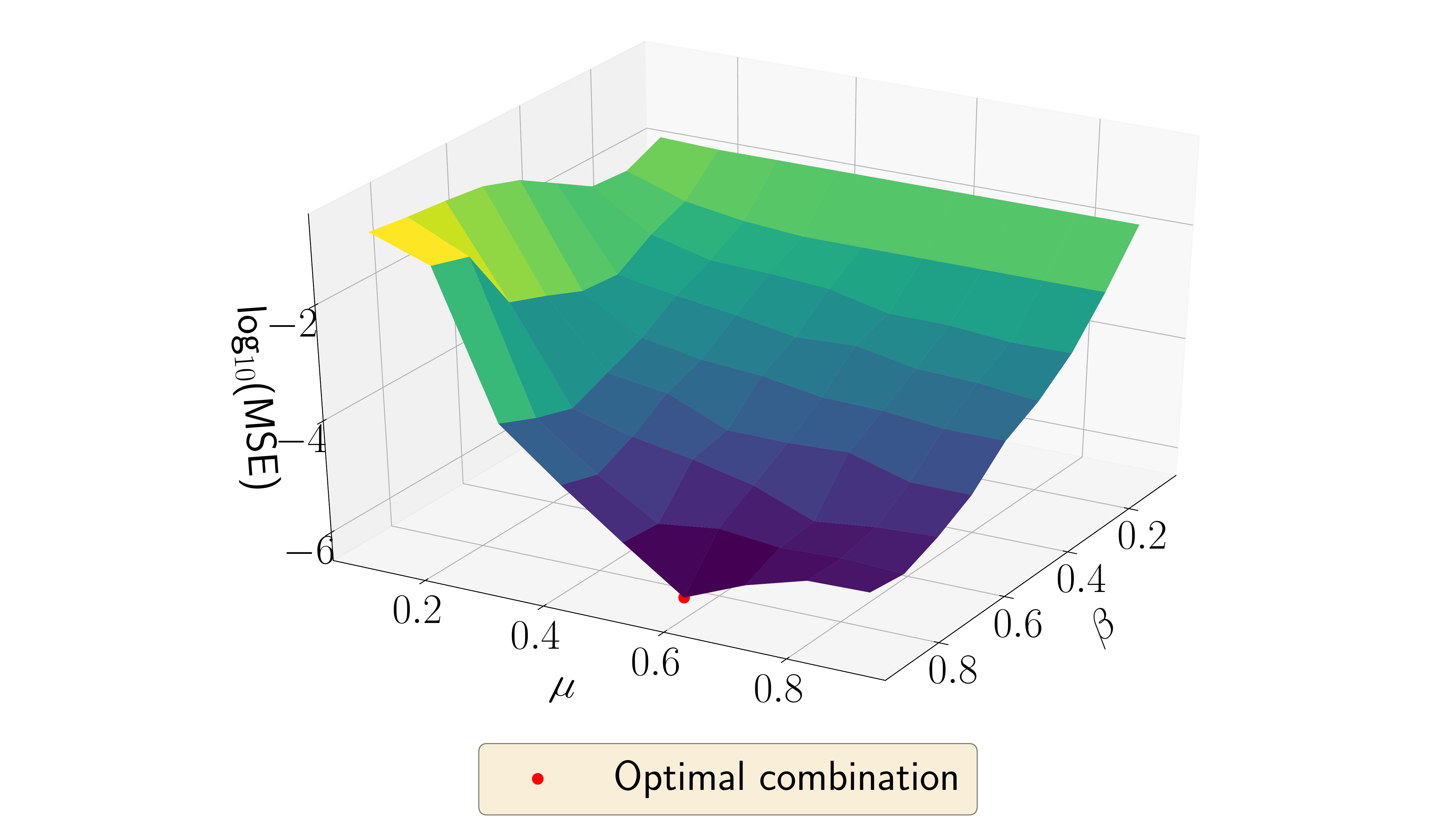}
    \caption{MSE values for different ($\mu, \beta$) combinations for $M=75$ arising from hyperparameter tuning for a Chebyshev-based formulation on an unstable 3D spiral ($c = 7$). }
    \label{fig:heatMap}
\end{figure}
\vspace{-3mm}
\section{Results}\label{section:results}
 
\noindent The performance of the proposed Chebyshev-based formulation of the latent space \ref{eq:latentSpaceSingleDim} for learning a DS using successive diffeomorphisms is presented in this section. This framework has been compared with the original FDM approach and the SDS-EF approach. The efficacy of using the proposed latent space has been evaluated by generating the forward orbits of the learnt dynamical systems from the original starting points. These forward orbits were compared with their respective original demonstrations. The replication generated by the learnt DS differed from the original demonstration in terms of speed, path traced and number of time-stamped points. Consequently, generally used methods such as the $L2$-norm could not be employed to measure the similarity between the demonstration and the replication. However, because both the replication and demonstration are temporal sequences, we employed FastDTW \cite{Salvador2004FastDTWTA}, a variant of dynamic time warping, to measure similarity. Lower scores are favourable. 
 

 
The following spiral trajectories (of unit radius and varying complexities) were used as demonstrations to evaluate and compare the performance of the different methods:
\begin{enumerate}
    \item Five 3-dimensional spherical spirals from an unstable DS
    \item Five 3-dimensional spherical spirals from a stable DS
\end{enumerate}
The formulation of the 3-dimensional spirals from an unstable DS (hereby referred to as unstable 3D spirals) and those from a stable DS (hereby referred to as stable 3D spirals) can be found in Appendix \ref{app6}. 
Test demonstrations were formulated using $c \in \{1,3,7,10,15\}$ for the unstable DS and $c \in \{1,2,3,5,7\}$ for the stable DS. From each of the forward orbits, $N = 500$ points were uniformly (temporally) sampled to be used for performance evaluation. As stated before, the value of $c$ regulates the complexity of the spiral. A spiral with higher $c$ makes the learning and replication tasks more challenging.
 
The tuned hyperaparameters utilised by the presented methods for learning the maps for unstable 3D spirals are presented in Tables \ref{tab:hParamsChebyshevPerrin} \& \ref{tab:hParamsSDSEF}.
 
\begin{table}
\centering
\caption{Hyperparameters associated with the proposed Chebyshev-based method and FDM for learning to replicate unstable 3D spirals}
\label{tab:hParamsChebyshevPerrin}
\begin{tabular}{|c|c|l|l|l|l|}
\hline
$\bm{c}$                    & \textbf{Method}                   & \multicolumn{1}{c|}{$\bm{\beta}$} & \multicolumn{1}{c|}{$\bm{\mu}$} & \multicolumn{1}{c|}{$\bm{M}$} & \multicolumn{1}{c|}{\textbf{MSE}} \\ \hline
                              & \cellcolor[HTML]{C0C0C0}Chebyshev & \cellcolor[HTML]{C0C0C0}0.9        & \cellcolor[HTML]{C0C0C0}0.6      & \cellcolor[HTML]{C0C0C0}50      & \cellcolor[HTML]{C0C0C0}7.38E-07  \\ \cline{2-6} 
\multirow{-2}{*}{\textbf{1}}  & FDM                               & 0.9                                & 0.8                              & 50                              & 2.52E-08                          \\ \hline
                              & \cellcolor[HTML]{C0C0C0}Chebyshev & \cellcolor[HTML]{C0C0C0}0.9        & \cellcolor[HTML]{C0C0C0}0.8      & \cellcolor[HTML]{C0C0C0}75      & \cellcolor[HTML]{C0C0C0}5.73E-07  \\ \cline{2-6} 
\multirow{-2}{*}{\textbf{3}}  & FDM                               & 0.9                                & 0.9                              & 75                              & 9.64E-08                          \\ \hline
                              & \cellcolor[HTML]{C0C0C0}Chebyshev & \cellcolor[HTML]{C0C0C0}0.5        & \cellcolor[HTML]{C0C0C0}0.9      & \cellcolor[HTML]{C0C0C0}175     & \cellcolor[HTML]{C0C0C0}7.26E-06  \\ \cline{2-6} 
\multirow{-2}{*}{\textbf{7}}  & FDM                               & 0.6                                & 0.9                              & 150                             & 3.35E-06                          \\ \hline
                              & \cellcolor[HTML]{C0C0C0}Chebyshev & \cellcolor[HTML]{C0C0C0}0.4        & \cellcolor[HTML]{C0C0C0}0.9      & \cellcolor[HTML]{C0C0C0}300     & \cellcolor[HTML]{C0C0C0}9.51E-06  \\ \cline{2-6} 
\multirow{-2}{*}{\textbf{10}} & FDM                               & 0.5                                & 0.9                              & 300                             & 5.39E-06                          \\ \hline
                              & \cellcolor[HTML]{C0C0C0}Chebyshev & \cellcolor[HTML]{C0C0C0}0.3        & \cellcolor[HTML]{C0C0C0}0.8      & \cellcolor[HTML]{C0C0C0}750     & \cellcolor[HTML]{C0C0C0}1.62E-05  \\ \cline{2-6} 
\multirow{-2}{*}{\textbf{15}} & FDM                               & 0.3                                & 0.8                              & 750                             & 1.56E-05                          \\ \hline
 
\end{tabular}
\end{table}

\begin{figure*}
     \centering
     \begin{subfigure}[b]{0.48\textwidth}
        \includegraphics[width=\linewidth,trim={8.5cm 0 11cm 0},clip]{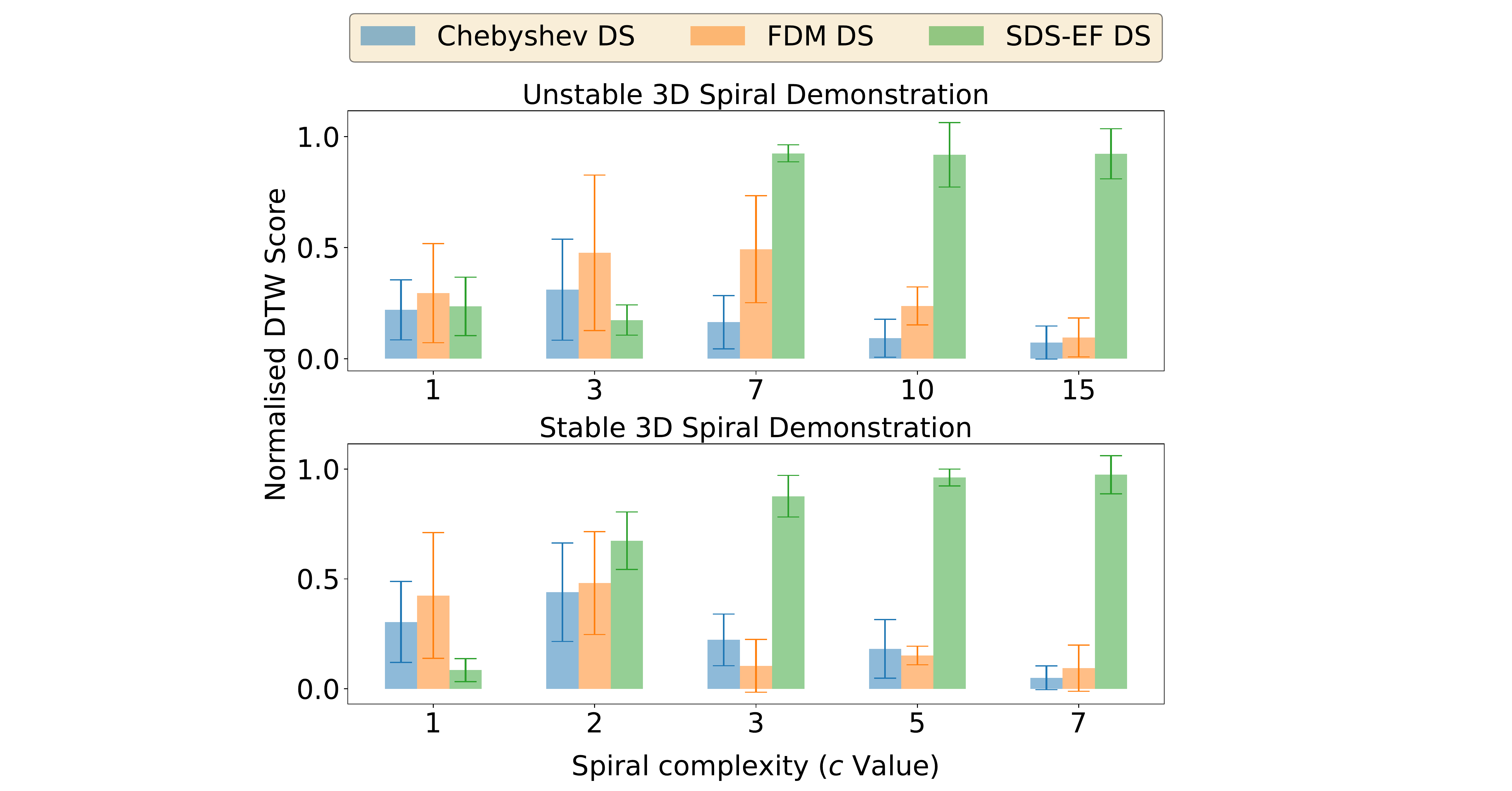}
        \caption{}
    \label{fig:dtwScoreRandom}
    
    \end{subfigure}
    \centering 
    \begin{subfigure}[b]{0.48\textwidth}
        \includegraphics[width=\linewidth,trim={9cm 0 11cm 0},clip]{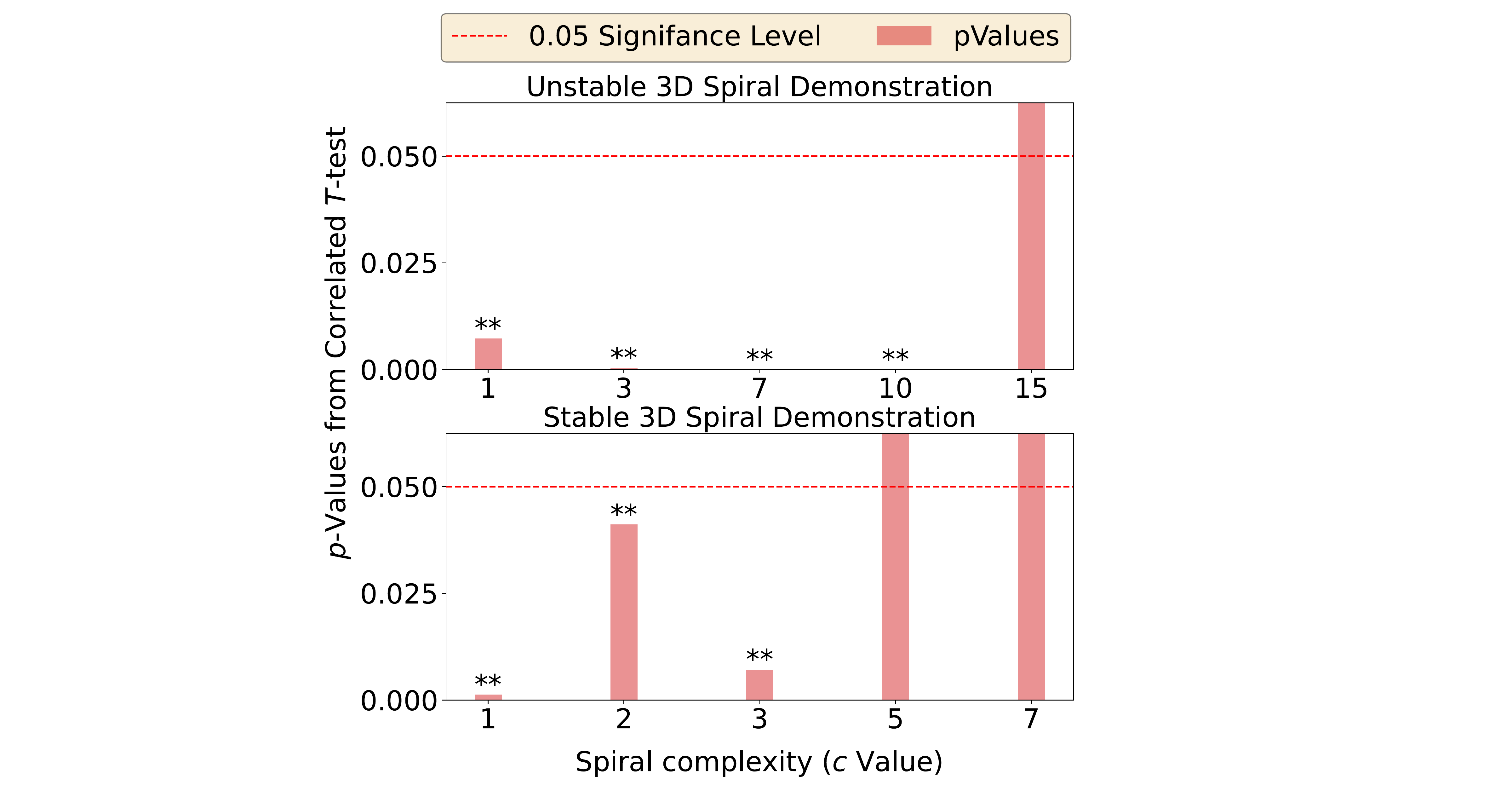}
        \caption{}
        \label{fig:ttestValues}
    \end{subfigure}
    
    \caption{Case-wise normalised similarity scores of replications generated from DS learnt using different methods -- proposed Chebyshev approach, FDM and SDS-EF -- for spirals of varying complexity. (left) Score computed using fast dynamic time warping (DTW) \cite{Salvador2004FastDTWTA} – a lower score is better; (right) $p-$values from the t-test conducted on the DTW scores of replications generated from DS learnt using the proposed Chebyshev approach and FDM for spirals of varying complexity}
\end{figure*}
 
\begin{figure*}
    \centering
    \includegraphics[width=0.8\textwidth, trim={5cm 3cm 4.5cm 0cm}, clip]{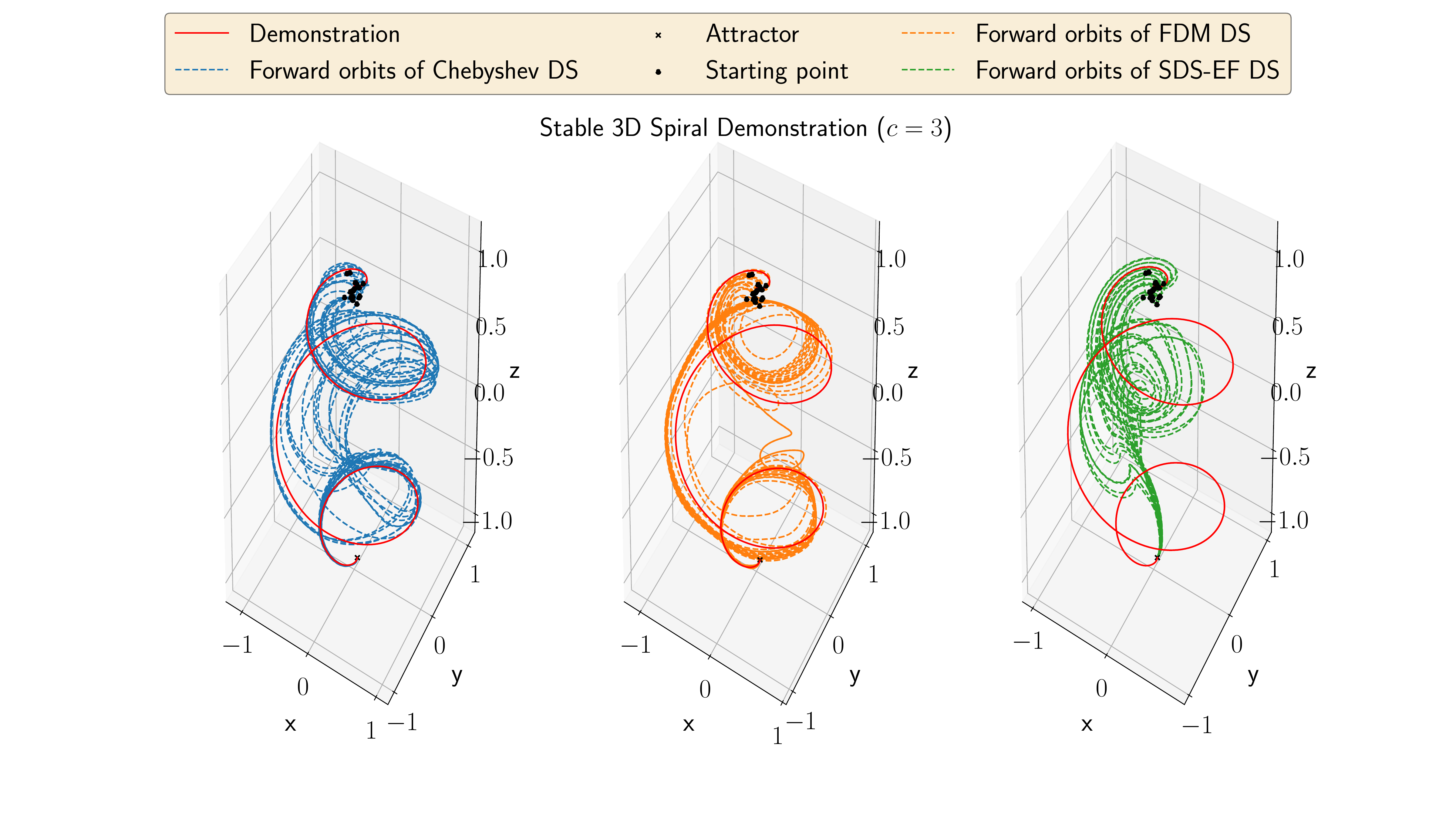}
        \caption{Replications generated by the DS learnt using different methods for stable 3D spirals ($c=3$)}
        \label{fig:stableReplications}
\end{figure*}

\begin{table}
\centering
\caption{Hyperparameters associated with the SDS-EF method for learning to replicate unstable 3D spirals}
\label{tab:hParamsSDSEF}
\begin{tabular}{|c|c|c|c|}
\hline
$\bm{c}$ & $\bm{L}$ & $\bm{n_f}$ & \textbf{MSE} \\ \hline
\textbf{1}          & 10              & 250               & 1.18E-02          \\ \hline
\textbf{3}          & 10              & 150               & 1.99E-02          \\ \hline
\textbf{7}          & 30              & 350               & 5.66E-02          \\ \hline
\textbf{10}         & 40              & 200               & 7.26E-02          \\ \hline
\textbf{15}         & 40              & 300               & 1.32E-01          \\ \hline
\end{tabular}
\end{table}

 \begin{figure*}
    \centering
    \includegraphics[width=0.75\linewidth, trim={10cm 0.5cm 12cm 0.5cm}]{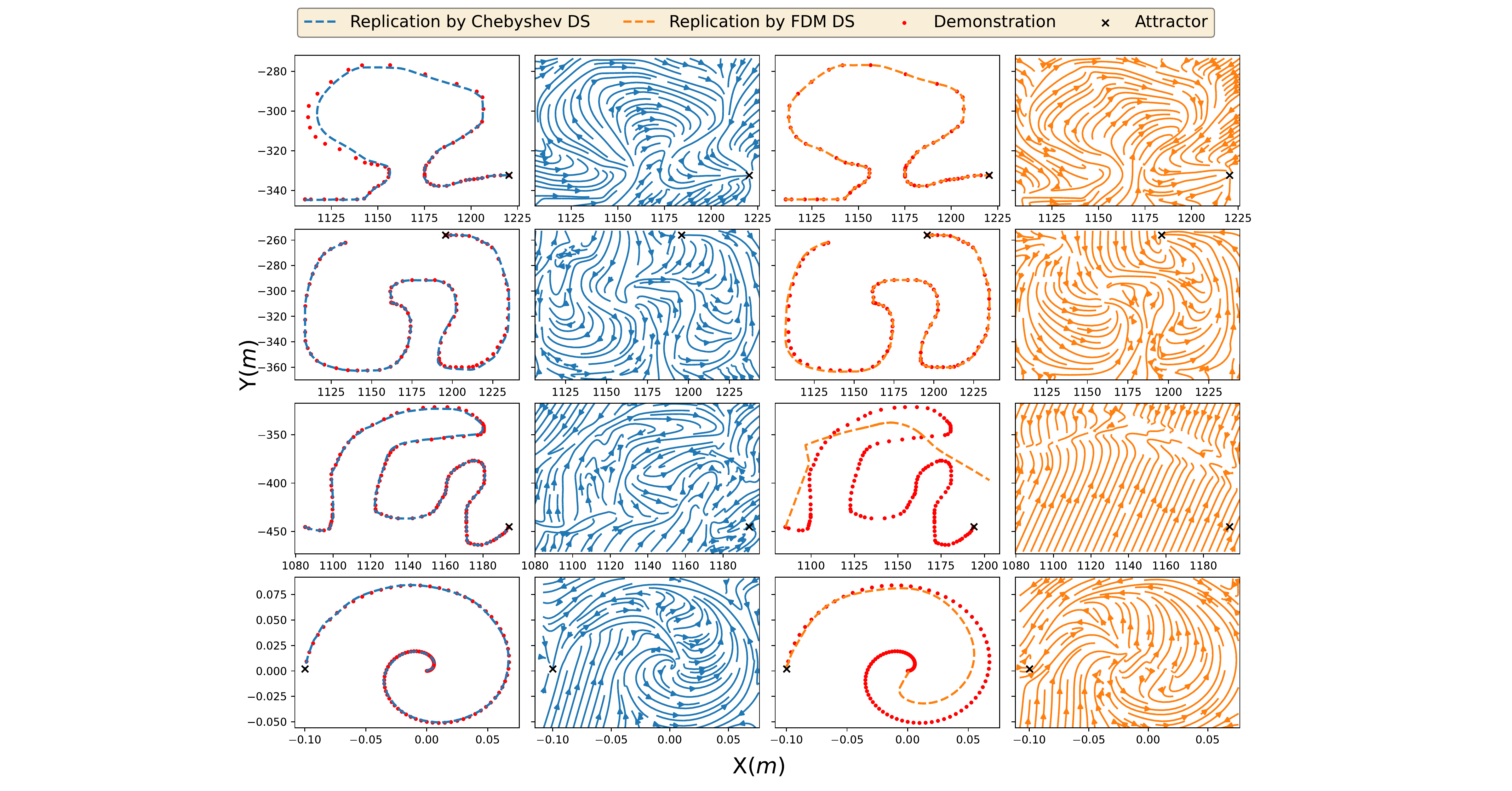}
    \caption{Replications generated by the DS learnt using the proposed Chebyshev-based method (first column) and the original FDM method (third column) for additional demonstrations; Vector field corresponding to the DS learnt using the Chebyshev-based method (second column) and the original FDM method (fourth column)}
    \label{fig:FDMvsChebyshev}
\end{figure*}

To generate the forward orbits of the learnt dynamical systems, 20 other starting points were randomly picked (consistent across different methods and spirals) in a vicinity of $0.1m$ around the original starting point $(0, 0, 1)$. The forward orbits generated for spiral cases $c=7$ (for unstable 3D spirals) and $c=3$ (for stable 3D spirals) are presented in Figures \ref{fig:replications} and \ref{fig:stableReplications}.

Each replication was compared to the original demonstration. The mean normalised DTW scores with their standard deviations are presented in Fig. \ref{fig:dtwScoreRandom}, where lower scores are favourable. Of the different methods, SDS-EF performs the best, closely followed by the proposed Chebyshev-based method when the spiral complexity is low. However, as the demonstration complexity increases, the proposed Chebyshev-based method takes the lead. For further clarification of cases where the performance of the proposed Chebyshev-based method and FDM seem close, the $p-$values from a comparative t-test are presented in \ref{fig:ttestValues}
 
The performances of the proposed approach and FDM on a 2-dimensional Archimedean spiral (presented in Appendix \ref{app6}) and on some hand-drawn characters are presented in \ref{fig:FDMvsChebyshev}. The replications show that the performance of the proposed method is on par with the FDM, and it also outperforms FDM: 
\begin{itemize}
    \item In the third case, the DS learnt using FDM converges at an incorrect attractor, consequently producing erroneous replication.
    \item In the fourth case, the DS learnt using FDM cannot capture the initial swirl of the spiral.
\end{itemize}    
 
\section{Robot Implementation}\label{section:robot_implementation}
\begin{figure}
  \centering
  \includegraphics[height=0.3\textheight, trim={35cm 40cm 15cm 40cm},clip]{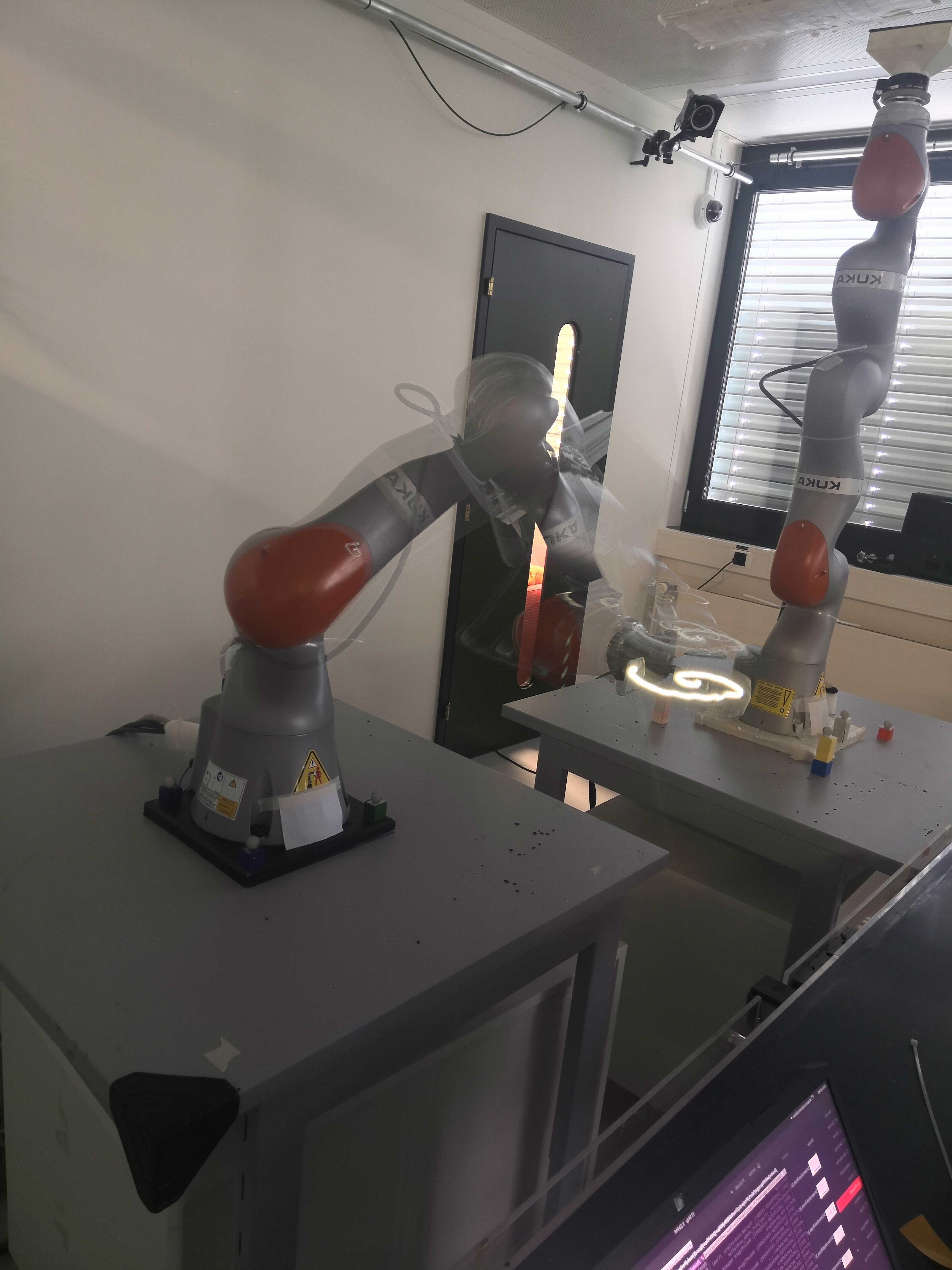}
\caption{Light painting of the replication of the 2D spiral recorded during the manipulator's motion}
    \label{fig:kukaReplication}
\end{figure}
\begin{figure*}
\centering
    \begin{subfigure}[b]{0.475\textwidth}
        \centering
        \includegraphics[width=0.45\linewidth, trim={30cm 0cm 30cm  0cm},clip]{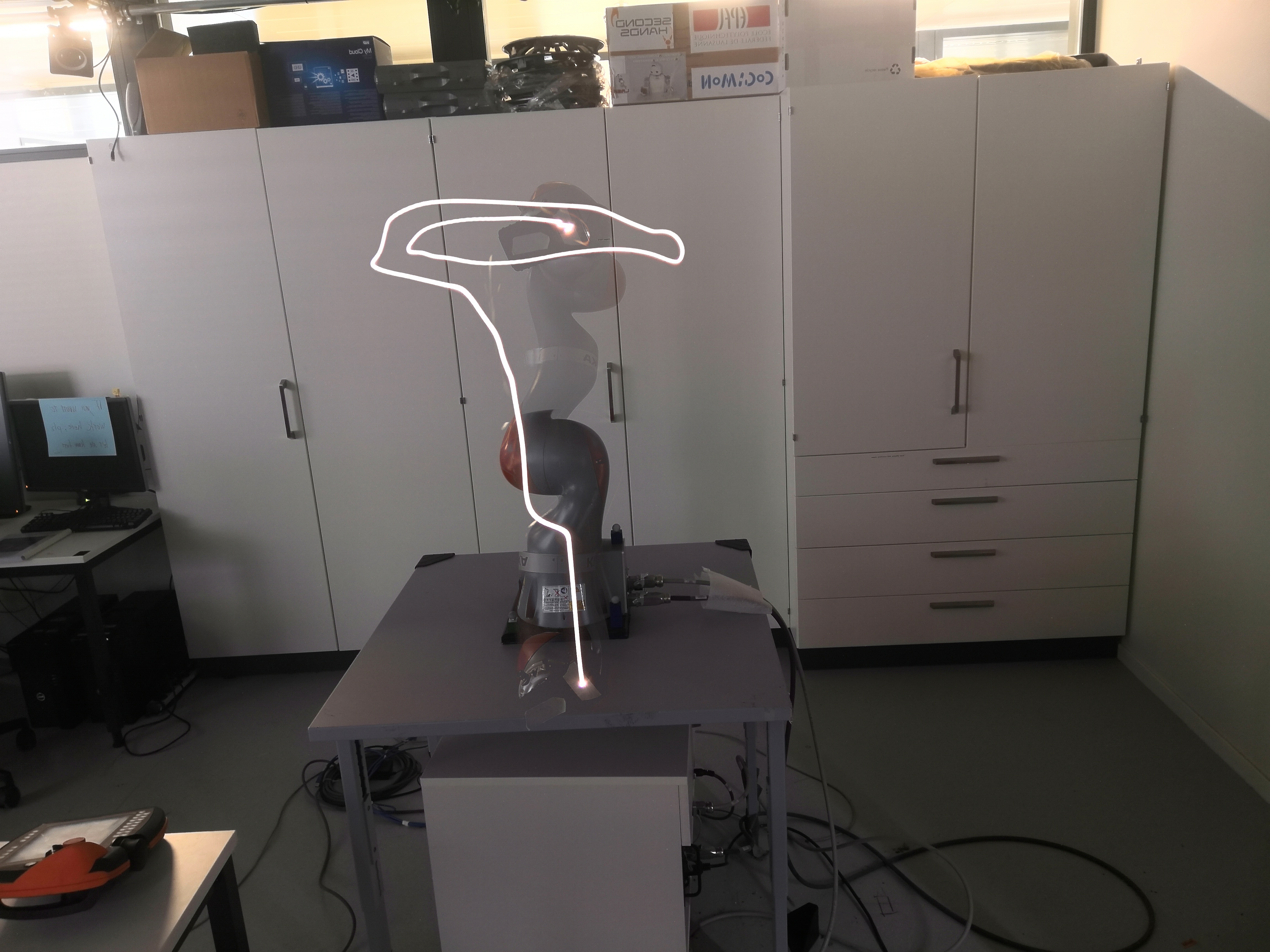}
        \includegraphics[width=0.525\linewidth, trim={10cm 0cm 10cm  0cm},clip]{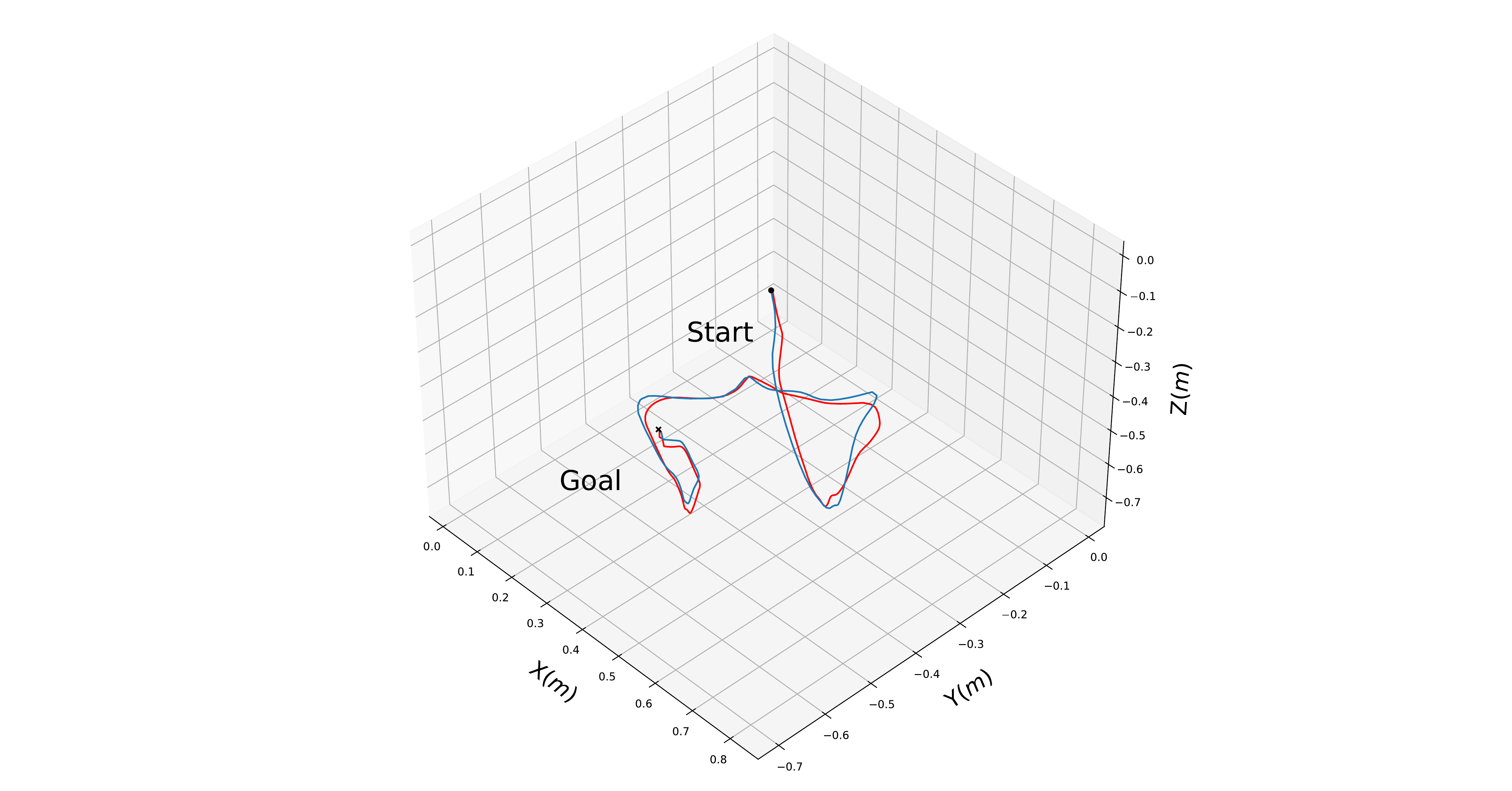}
        \caption{}
    \end{subfigure}
    \begin{subfigure}[b]{0.475\textwidth}
        \centering
        \includegraphics[width=0.45\linewidth, trim={10cm 0cm 50cm  0cm},clip]{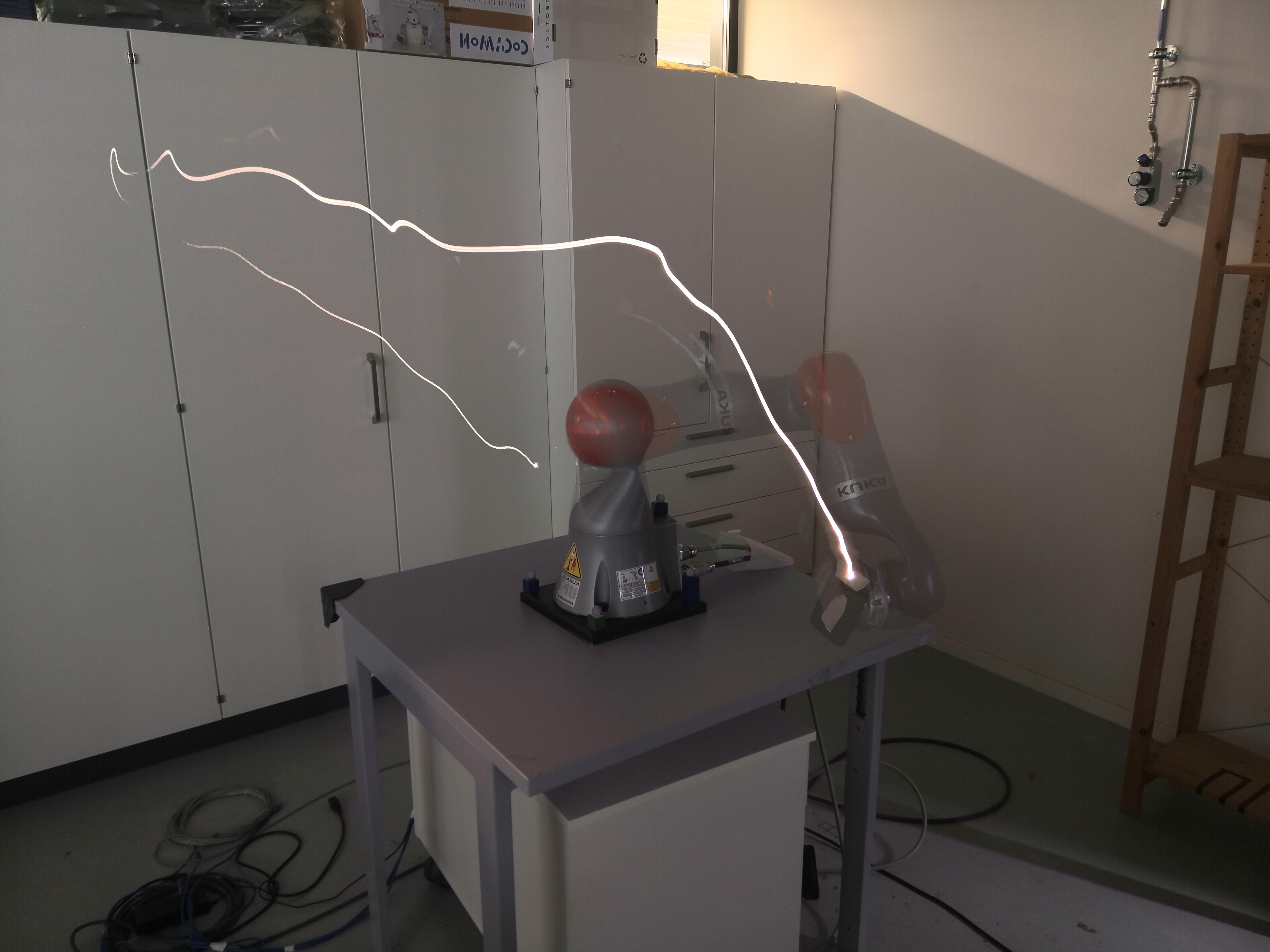}
        \includegraphics[width=0.525\linewidth, trim={10cm 0cm 10cm  0cm},clip]{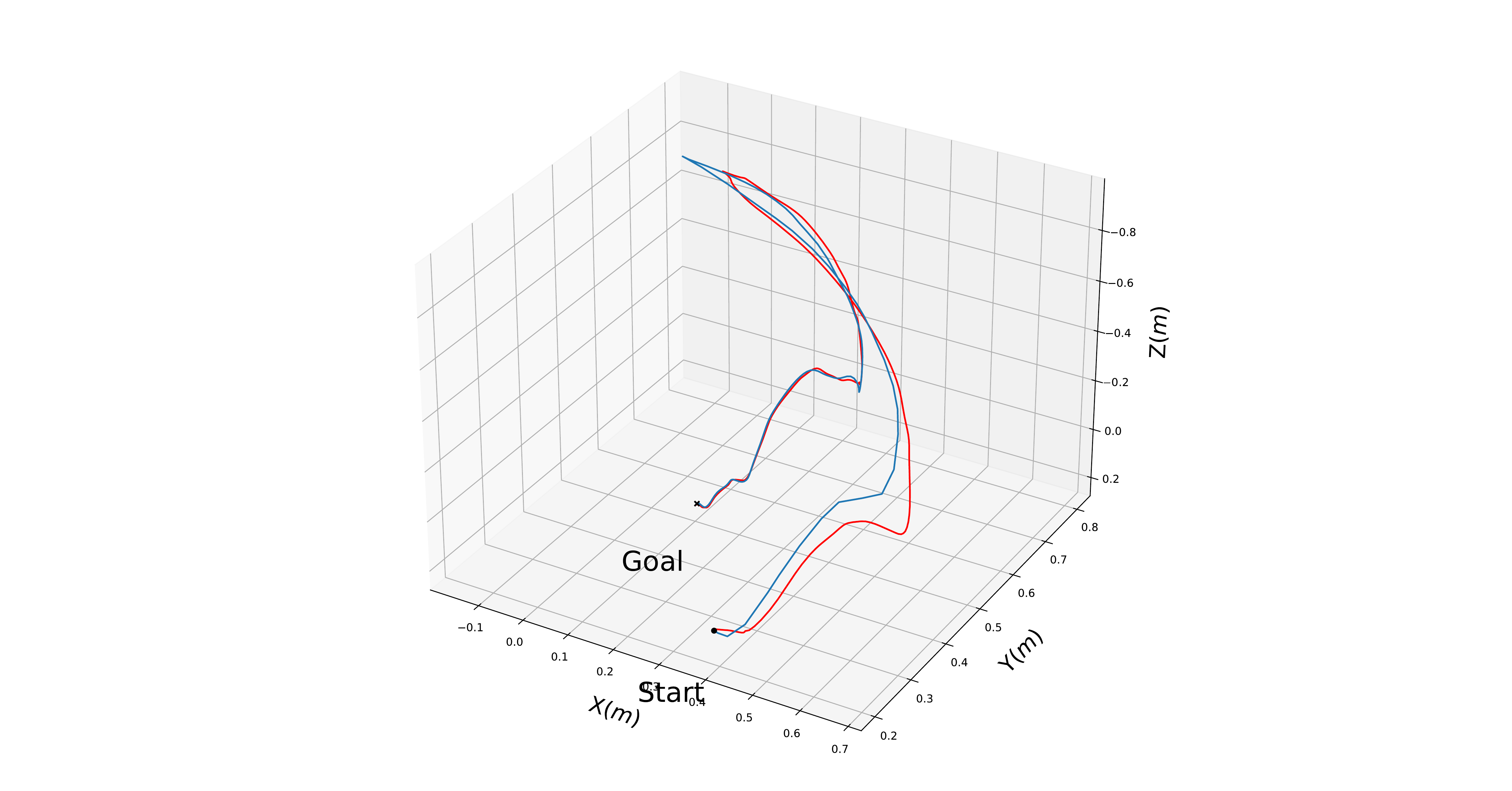}
        \caption{}
    \end{subfigure}
    \hfill
    \begin{subfigure}[b]{0.475\textwidth}
        \centering
        \includegraphics[width=0.45\linewidth, trim={30cm 0cm 30cm  0cm},clip]{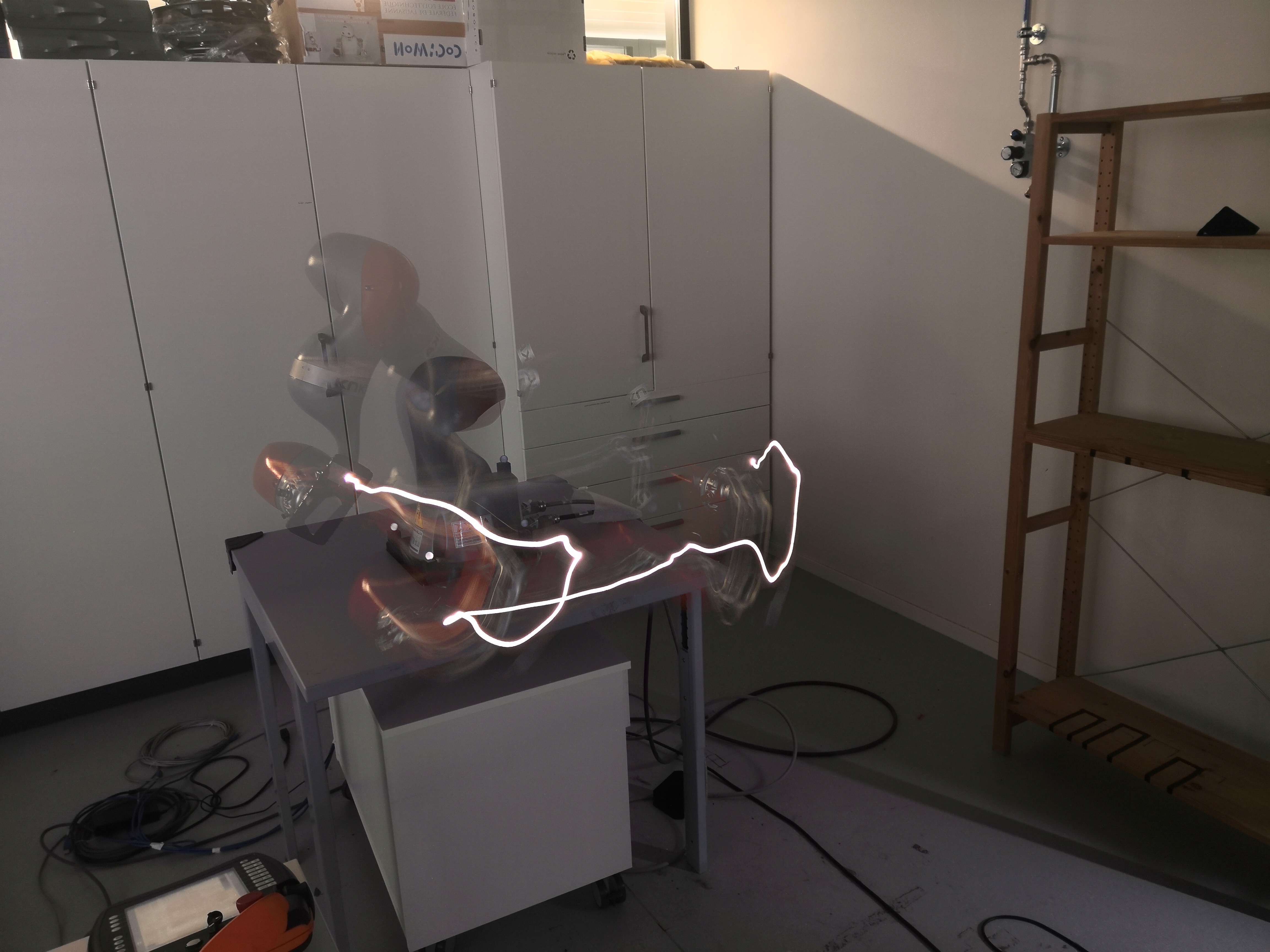}
        \includegraphics[width=0.525\linewidth, trim={10cm 0cm 10cm  0cm},clip]{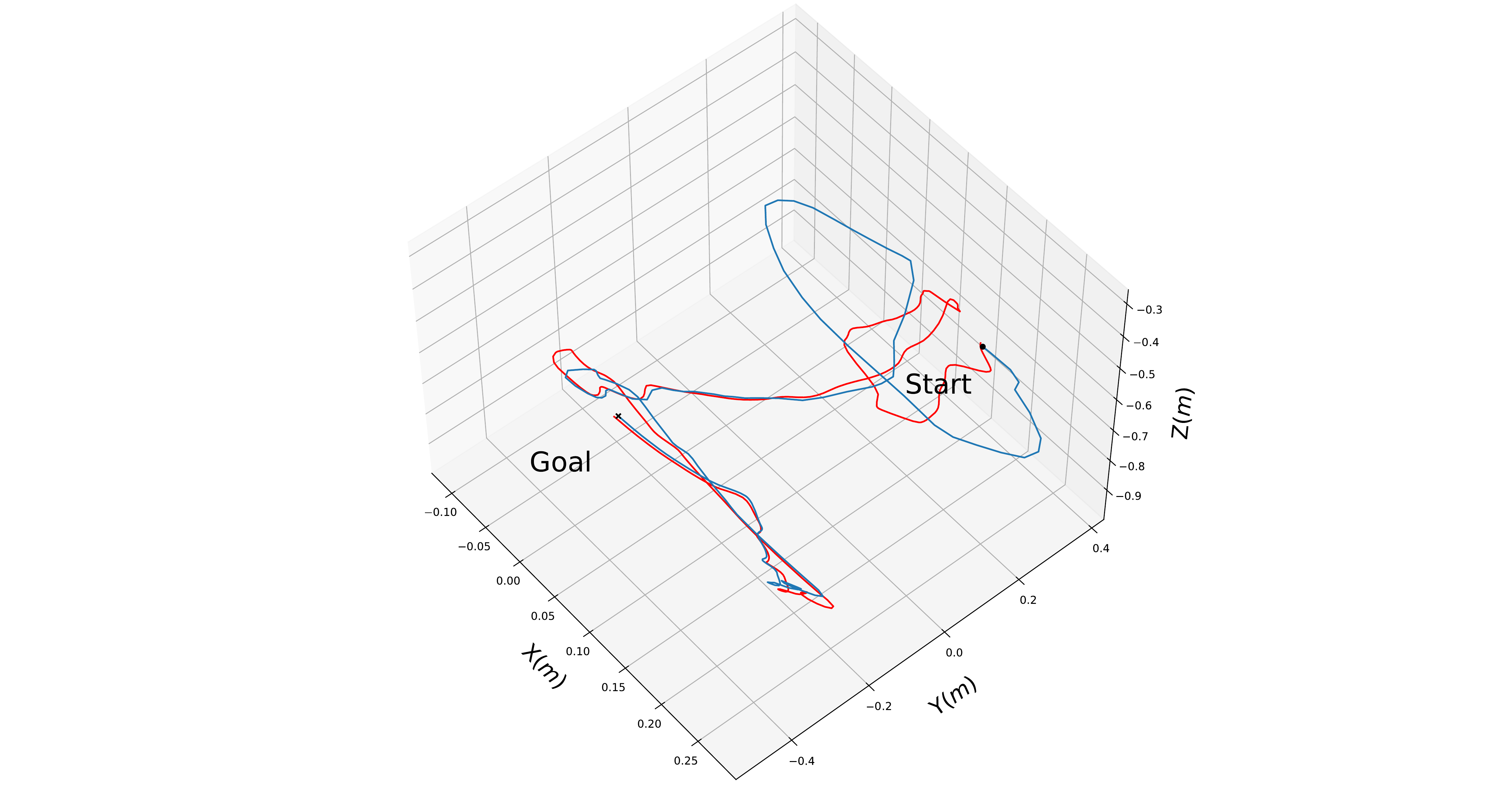}
        \caption{}
    \end{subfigure}
    \begin{subfigure}[b]{0.475\textwidth}
        \centering
        \includegraphics[width=0.45\linewidth, trim={20cm 0cm 40cm  0cm},clip]{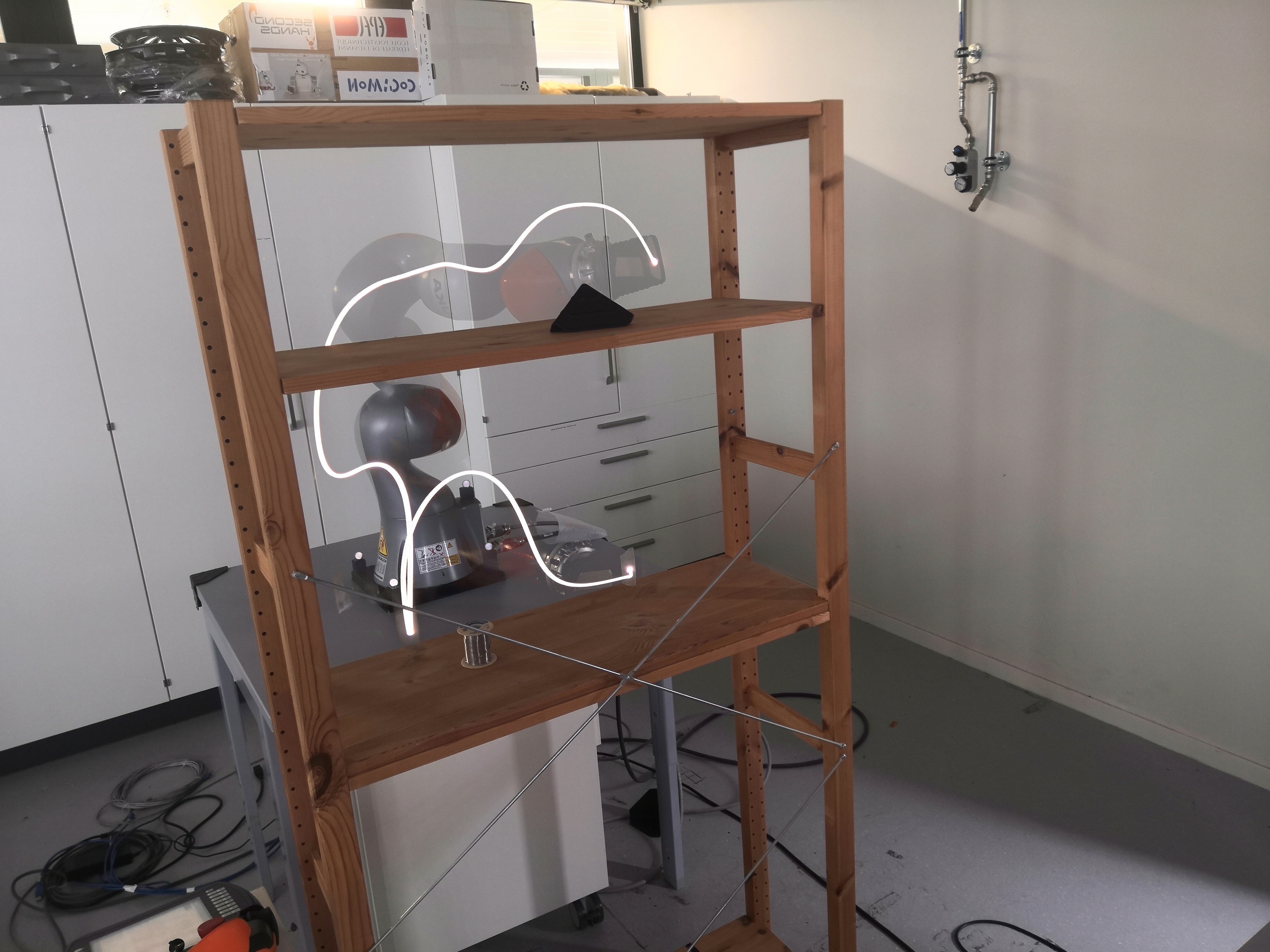}
        \includegraphics[width=0.525\linewidth, trim={10cm 0cm 10cm  0cm},clip]{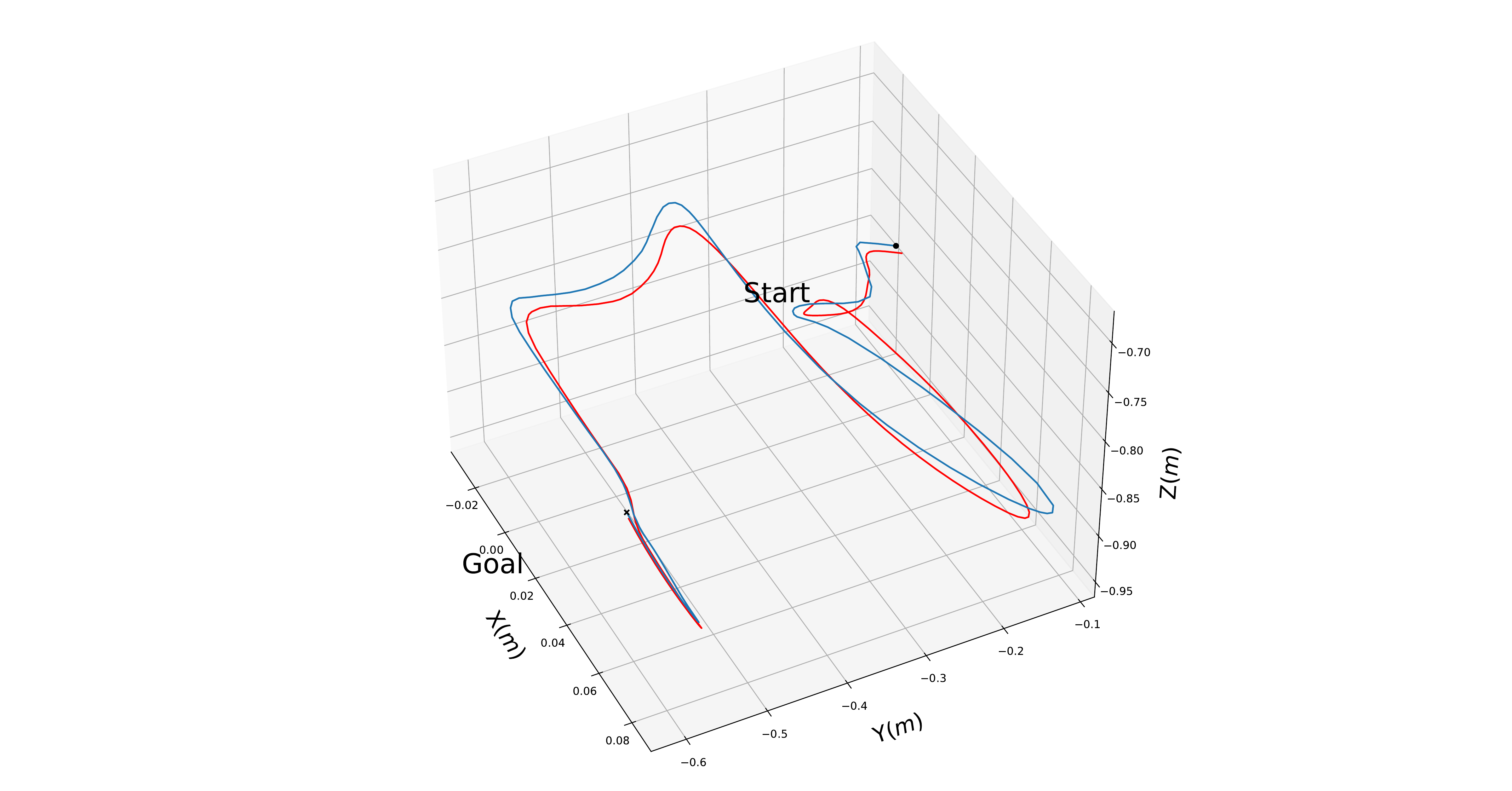}
        \caption{}
    \end{subfigure}
    \hfill
    \begin{subfigure}[b]{0.475\textwidth}
        \centering
        \includegraphics[width=0.45\linewidth, trim={0cm 10cm 0cm  10cm},clip]{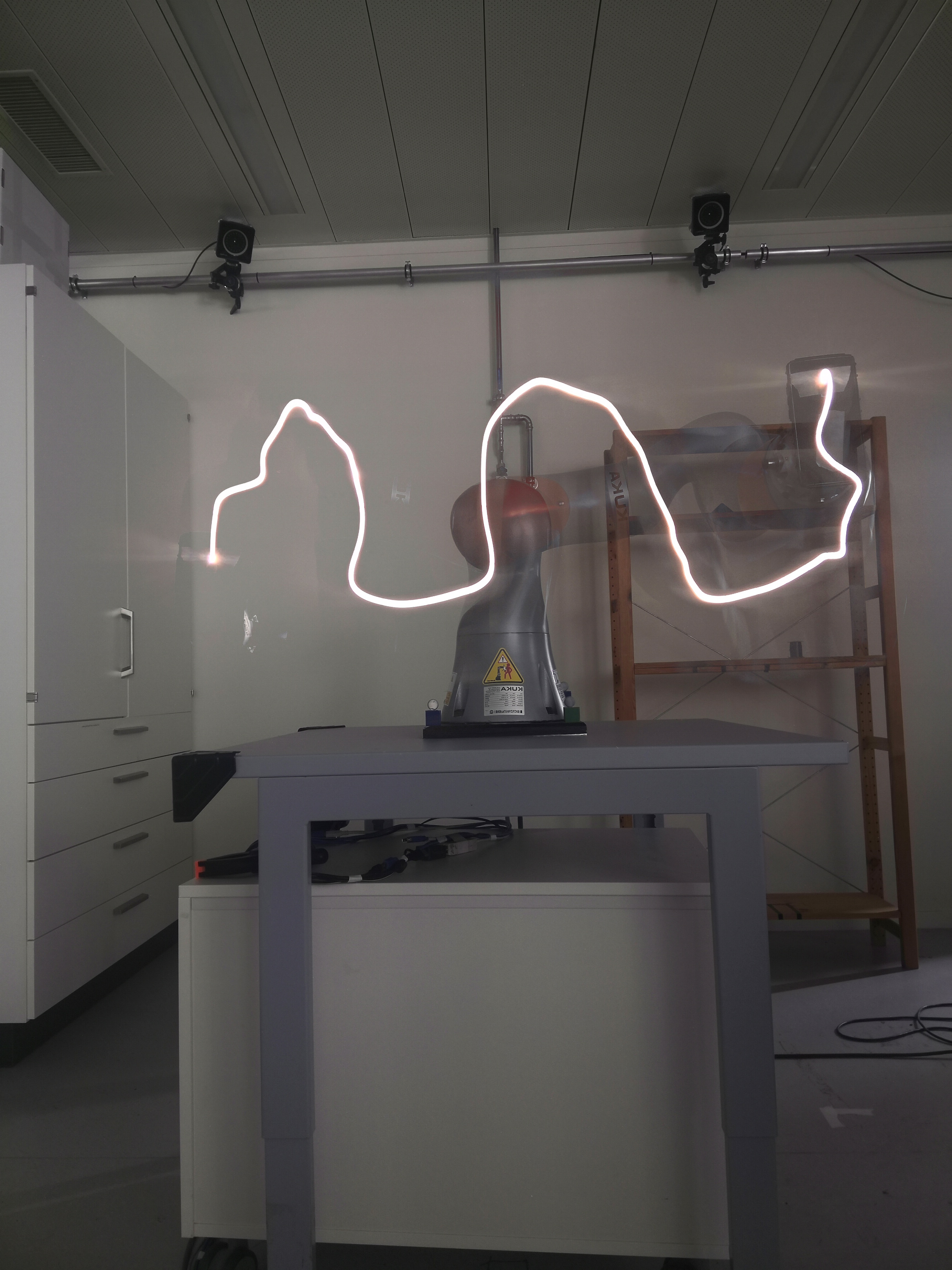}
        \includegraphics[width=0.525\linewidth, trim={10cm 0cm 10cm  0cm},clip]{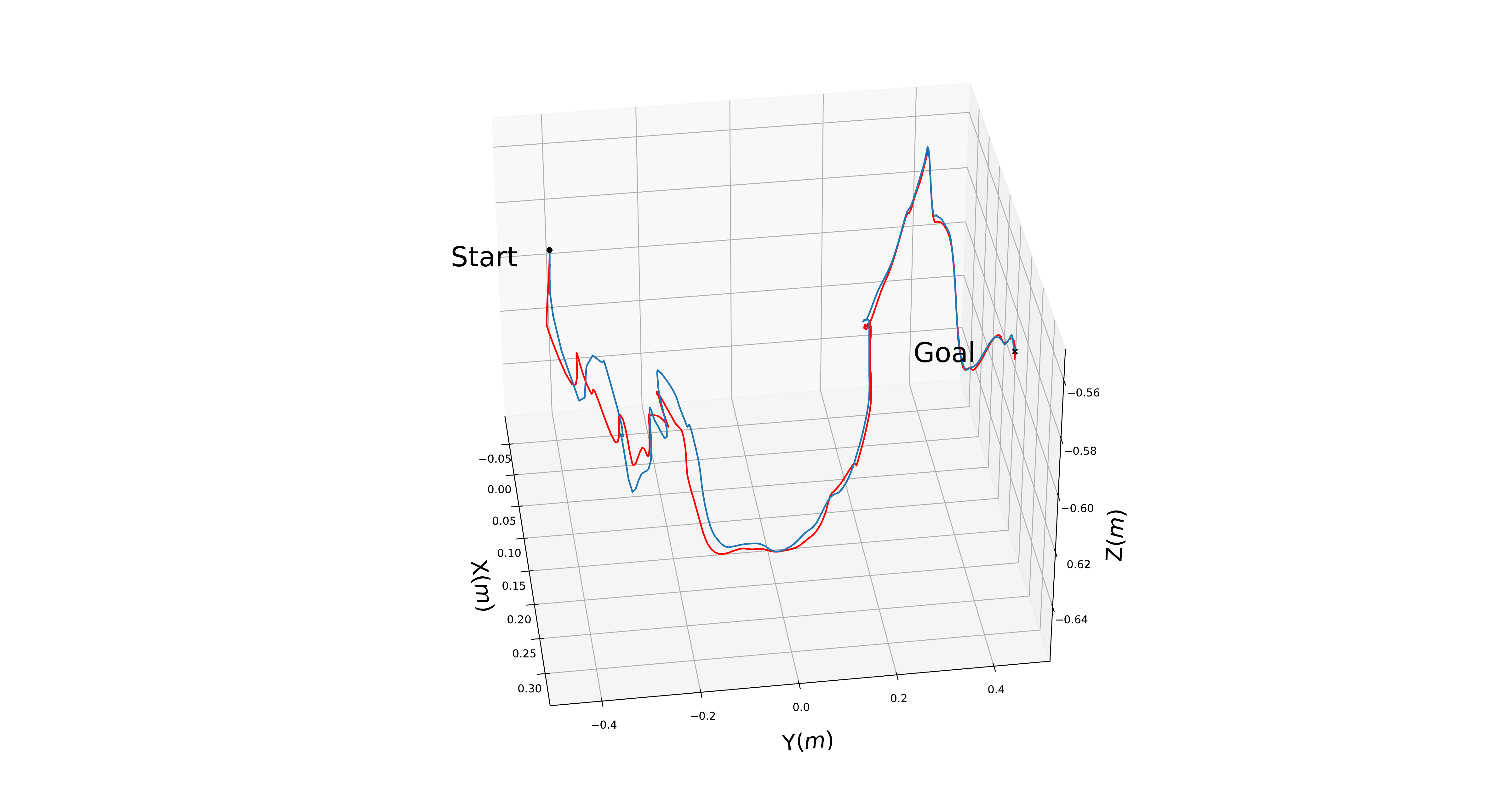}
        \caption{}
    \end{subfigure}
    \begin{subfigure}[b]{0.475\textwidth}
        \centering
        \includegraphics[width=0.45\linewidth, trim={0cm 10cm 0cm  10cm},clip]{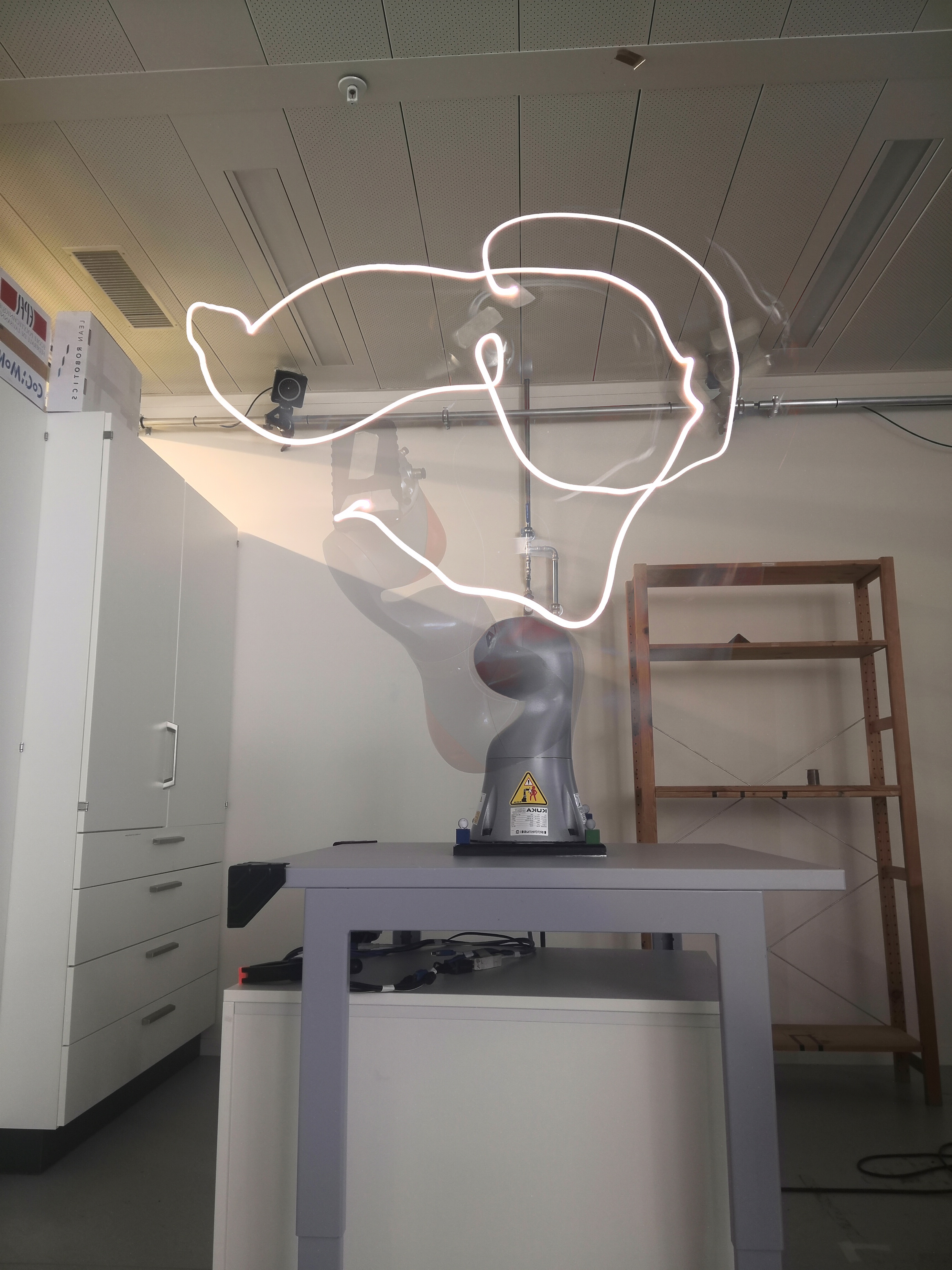}
        \includegraphics[width=0.525\linewidth, trim={10cm 0cm 10cm  0cm},clip]{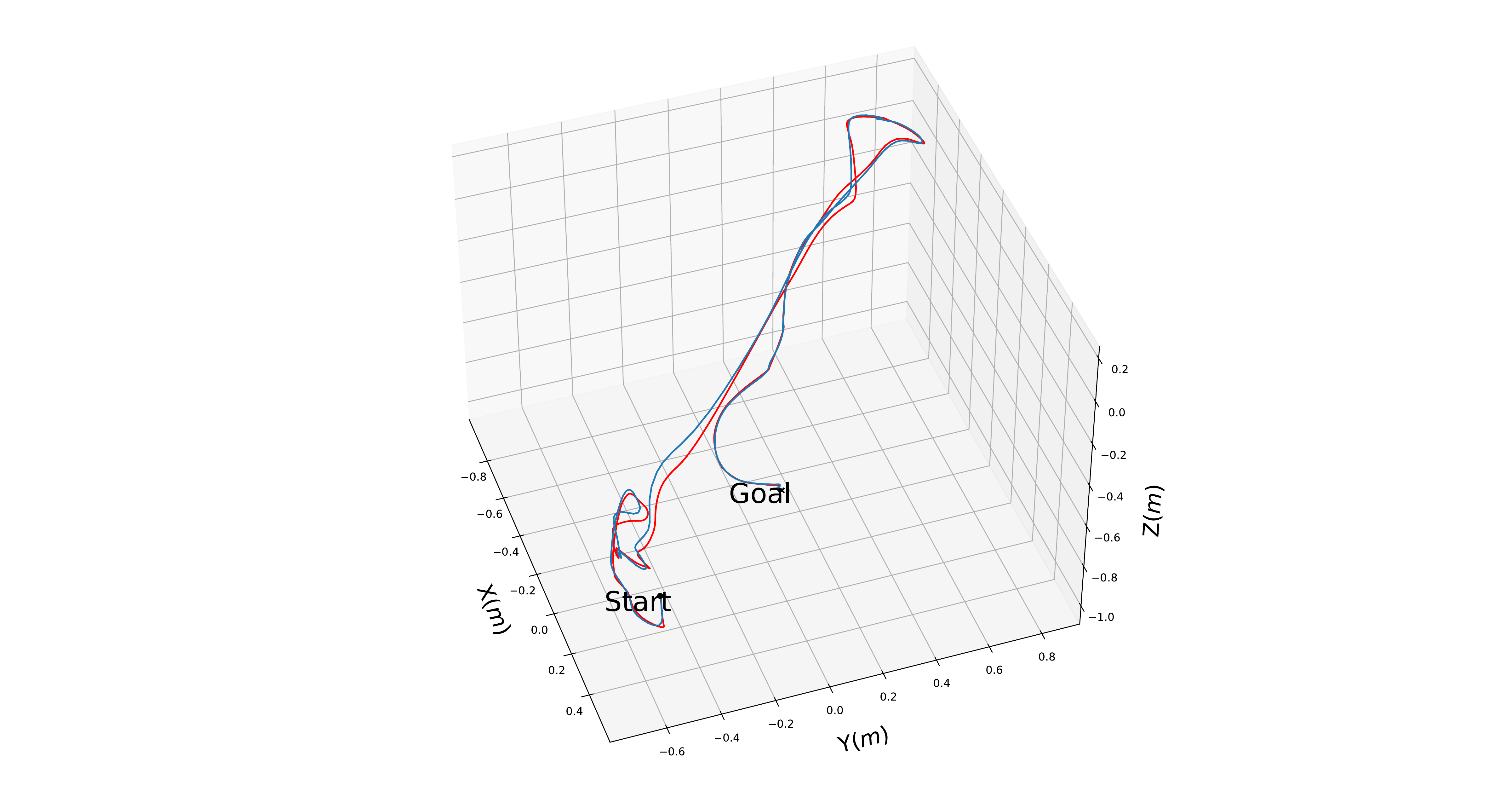}
        \caption{}
    \end{subfigure}
    \hfill
    \caption{Cases (a) through (f): Pairs of light painting of the kinesthetically supplied demonstrations and comparison between the Cartesian position of the end-effector during the demonstration (red) and the replication (blue) produced by the learnt DS}
    \label{fig:kinestheticDemos}
\end{figure*}
 
\begin{table}[h]
\centering
\caption{Similarity scores between the kinesthetic demonstration and the corresponding replication. A lower score is better}
\label{tab:dtwKinesthetic}
\begin{tabular}{|c|c|c|c|c|c|c|}
\hline
\textbf{Case}                                                & a     & b     & c     & d    & e    & f     \\ \hline
\textbf{\begin{tabular}[c]{@{}c@{}}Score\end{tabular}} & 25.92 & 19.28 & 81.25 & 9.52 & 9.74 & 18.31 \\ \hline
\end{tabular}
\end{table}
\noindent Eight demonstrations of varying dimensionality and complexity were provided to the robot as part of the robot implementation. These were as follows:
\begin{itemize}
    \item Numerically generated demonstrations (2):
    \begin{itemize}
        \item 2-dimensional task-space data corresponding to the end-effector's motion on an Archimedean 2D spiral with a radius of $15$ cm (similar to \ref{eq:archimideanSpiral})
        \item 3-dimensional task-space data corresponding to the end-effector's motion on an unstable 3D spiral ($c = 7$) with a radius of $20$ cm
    \end{itemize}
    \item Kinesthetically recorded demonstrations (6): 7-dimensional joint-space data corresponding to four user-supplied demonstrations
\end{itemize}
The DSs corresponding to them were learnt using the proposed Chebyshev-based latent space formulation in conjunction with the FDM approach (for diffeomorphism learning) from a single demonstration. The learnt DSs were next used to generate replications on a KUKA IIWA LWR 14. For the two task-space demonstrations, the Cartesian position of the end-effector yielded by the learnt DS was supplemented with a constant 3-dimensional orientation vector. The 6-dimensional end-effector pose was commanded to the robot via a controller based on \cite{kronander_passive_2016}. The light paintings of the replications generated by the manipulator, which were generated using a light source mounted on the manipulator's end-effector, are presented in Fig. \ref{fig:kukaReplication}. For the six kinesthetic demonstrations, the 7-dimensional joint position commanded by the learnt DS was directly published on the robot. The light paintings corresponding to the demonstrations and the comparison with their replications are presented in Fig. \ref{fig:kinestheticDemos}. The similarity scores computed using FastDTW are presented in Table \ref{tab:dtwKinesthetic}.
 
\section{Conclusion}\label{section:conclusions}
\noindent In this article, a latent space representation of a single demonstration was obtained. The representation was quasi-linear, given a demonstration of a $2$-dimensional or higher Euclidean space. The theoretical findings were verified with numerical experiments. Lower DTW scores implying higher replication accuracy were obtained for several demonstrations in which the proposed latent space was used instead of the commonly chosen Hurwitz matrices to describe linear dynamics in the latent space.
 
However, the proposed latent space does not affect the learnt diffeomorphism. In the present work, a state-of-the-art method of diffeomorphic point matching (FDM) is employed to learn the diffeomorphism.
In the future, we would like to incorporate the knowledge of the latent space to simplify the learning of the diffeomorphism.
 
\begin{appendices}
\section{}\label{app3}
\textbf{Proof of Lemma \ref{Lemma2}}
\newline
Following Proposition \ref{propn3}, the eigenvector entries $u_i$ of the eigenvector $u =[u_1, \dotsc, u_N]$ are
\begin{align}
    & u_{i}(\lambda) = u_{0}\left[T_i(1-\frac{\lambda}{2}) -  \frac{\lambda}{2}V_{i-1}\:(1-\frac{\lambda}{2})\right]
\end{align}
for an arbitrary $u_0\in \R$ and with $T_i$ and $V_i$ defined as
\begin{align}
    &T_i(\theta) = \operatorname{cos}(i \operatorname{arccos}(\theta)), \quad V_i(\theta) = \frac{\operatorname{sin}((i+1) \operatorname{arccos}(\theta))}{\operatorname{sin}(\operatorname{arccos}(\theta))}\label{tivi}
\end{align}
 
To study of the stationary points of Eq.~\ref{eq:chebyshevFormulation}, we differentiate Eq.~\ref{eq:chebyshevFormulation} with respect to the index $i$:
 
\begin{align}
    \ddi {u_i} & = u_0 \left[ - \theta \sin((i-1) \theta)  - \theta \frac{\lambda}{2} \frac{\cos((i-1) \theta)}{\sin(\theta)}\right], \\
    \theta & \coloneqq \cos^{-1}{(1-\frac{\lambda}{2})} \label{eqn:chebyshev_derivative}
\end{align}
Therefore, the series $\{ u_i\}$ is either monotonically increasing or monotonically decreasing for $\lambda$ and $i$ such that 
\begin{equation}\label{mon_cond}
    \sin((i-1) \theta)  + \frac{\lambda}{2} \frac{\cos((i-1) \theta)}{\sin(\theta)} = 0
\end{equation}
Let the eigenvalue $\lambda\in [0,4]$ be expressed as
\begin{equation}
    \lambda = 2 \left( 1 - \cos(\theta - 2 \pi j) \right), \quad j \in \mathbb{N}. 
\end{equation}
Replacing in \eqref{mon_cond}, we obtain
\begin{align}
    \tan((i-1) \theta) + \tan(\frac{\theta}{2} - \pi j)                                                         & = 0.
\end{align}
Therefore, the stationary points correspond to all $i$, such that
\begin{equation}\label{eqn:upper_monotonicity}
    i = \frac{\pi j}{\theta} + \frac{1}{2}=\frac{\pi j}{\cos^{-1}(1-\frac{\lambda}{2})} + \frac{1}{2}
\end{equation}
As \eqref{eqn:upper_monotonicity} must hold for $i=N$, $\lambda$ should satisfy \eqref{mon_cond_lambda}
 
\section{}\label{app4}
\textbf{Proof of Lemma \ref{Lemma3}}
\newline
Define
\begin{align}
   &B_i = B_1+ \alpha(i) B_2, \quad \alpha(i) = 2 \cos \left( \frac{2\pi i}{K} \right)\label{Bformat},\\
   & i= \begin{dcases} 1,2,\dotsc,\frac{K-1}{2} \quad \text{$K$ is odd} \\
   1,2,\dotsc,\frac{K}{2}-1 \quad \text{$K$ is even}
  \end{dcases}\nonumber
\end{align}
 
Observe that the smallest repeating eigenvalue of $L(G)$ is the largest repeating eigenvalue of $J$. From Proposition \ref{propn2}, this is the largest eigenvalue in one of the $B_i$ matrices defined in \eqref{Bformat}.
  Define
  \begin{align}\label{Mdef}
          & B_i= I+M_i, \\
          & M_i= \begin{pmatrix}
         0 & 1 & 0 & \cdots &0 \\
         1 & -1 & 1& \cdots &0 \\
         \vdots &\vdots &\vdots &\ddots &\vdots\\
         0 &0 & &0 \cdots & \alpha(i)-2
       \end{pmatrix}= M_1(i)+M_2,
\end{align}
with $\alpha(i)$ defined in \eqref{Bformat}. The matrices $M_1(i)$ and $M_2$ are defined as 
       \begin{align*} & M_1(i) = \begin{pmatrix}
                                      M_0 & v \\
           v^\top &  \alpha(i)-2
                                   \end{pmatrix}, \quad M_2= diag(0, -1,\cdots, -1,0 )
\end{align*}
where , and
\[ M_0= \begin{pmatrix}
                                     0 & 1  &\cdots &0 &0 \\
                                     1 & 0  &\cdots &0 &0\\
                                     \vdots &\vdots  &\ddots &\vdots &\vdots\\
                                     0 &0 &\cdots &0 & 1\\
                                     0 &0 &\cdots &1 & 0
                                   \end{pmatrix}\in \R^{(N-1)\times (N-1)}, \quad v= \begin{pmatrix}
                                                             0 \\
                                                             \vdots \\
                                                             1
                                                           \end{pmatrix}
\]
 
From Theorem 3.7 in \cite{bapat2010graphs}, because $M_0$ is the adjacent matrix of a path graph with $N-1$ vertices, and therefore,
\[ Eig(M_0) = 2 cos\left( \frac{\pi j}{N}\right), \quad j =1,\dotsc,N-1.
\]
Let $\lambda_N$ and $\lambda_1$ denote the largest and smallest eigenvalues of an $N \times N$ matrix. From Theorem 4.3.17 in \cite{horn2012matrix}, the largest eigenvalue of each $M_1(i)$ matrix is bounded below by
\[2 cos\left( \frac{\pi}{N}\right)=\lambda_{N-1}(M_0) \leq \lambda_{N}(M_1(i)), \quad \text{for all $i$}
\]
From Corollary 4.3.15 in \cite{horn2012matrix}, the largest eigenvalue of all $M_i$ matrices is bounded above and below by
\begin{align} 
2 cos\left( \frac{\pi}{N}\right) -1=& \lambda_N(M_1(i))+ \lambda_1(M_2)\nn\\
&< \lambda_N(M_i)<2 cos\left( \frac{\pi}{N}\right) \quad \text{for all $i$}\label{bounds}
\end{align}
The inequality is strict as $M_1(i)$ and $M_2$ do not have a common eigenvector. From \eqref{Mdef},
\[ Eig(L(G))=1-Eig(M_i), \quad i = \begin{dcases} 1,2,\dotsc,\frac{K-1}{2} \quad \text{$K$ odd}\\
   1,2,\dotsc,\frac{K}{2}-1 \quad \text{$K$ even}
  \end{dcases}
\]
As the eigenvalues of all $M_i$ repeat with multiplicity equal to $2$, therefore, from \eqref{bounds}, the smallest $K-1$ repeating eigenvalues of $L$ (assuming $K$ is odd) are upper bounded as given by \eqref{small_eig_L_bds}.
 
\section{}\label{app5}
\textbf{Proof of Proposition \ref{propn5}}
\newline

The $n$ smallest repeating eigenvalues belong to the order $ \Oh(1/N^2)$ from \eqref{small_eig_L}. The eigenvectors of $L(G)$ under consideration are same as those of $B_i$ and satisfy its eigenequation, therefore,
  \begin{align}
      &B_1 u^i + \alpha(i) B_2 u^i = \lambda_i u^i, \quad \alpha(i)= 2\cos \left(\frac{2\pi i}{n+1}\right),\nn \\
      &i = \begin{dcases} 1,2,\dotsc,\frac{n}{2} \quad \text{$n$ is even}\\
   1,2,\dotsc,\frac{n-1}{2} \quad \text{$n$ is odd}
  \end{dcases}\label{eig_eqn_component}
  \end{align} 
  
We suppress the dependency on $i$ for simplicity and denote $u^i =[u_1^i,\dotsc, u_N^i]$. Consider $\norm{u_1^i }_2 \in \Oh(1)$ to be fixed for all $i$. Expanding the eigenequation \eqref{eig_eqn_component} we obtain
  \begin{align*}
   u_2^i= & \lambda_i u_1^i- u_1 = -u_1+ \Oh \parb{1/N^2}\\
    u_3^i =&\lambda_i u_2^i- u_1 = - u_1+ \Oh \parb{1/N^2} \\
    u_4^i = &\lambda_i u_3^i- u_2^i =  u_1+ \Oh \parb{1/N^2}\\
    \vdots&\\
    u_N^i= & \frac{1}{1-\lambda_i} u_{N-1}^i= \pm u_1+\Oh \parb{1/N^2}
  \end{align*}
This shows that $\norm{u^l -u^m}_2 \in  \Oh\parb{1/N^2}$, wherein $u^l$ and $u^m$ denote the first $N$ components of two arbitrary vectors in the set of eigenvectors of $L(G)$ corresponding to the smallest $n$ repeating eigenvalues. 

\section{}\label{app6}
\textbf{Unstable 3D spirals}
\begin{align}
    \begin{split}
        & x_t = \operatorname{sin}(t)\operatorname{cos}(ct)\\
        & y_t = \operatorname{sin}(t)\operatorname{sin}(ct)\\
        & z_t = \operatorname{cos}(t)
    \end{split}
    \label{eq:spiralDemonstration}
\end{align}
for $t \in [0, 3.14]$. The constant term $c$ regulates the complexity of the spiral.
 
\textbf{Stable 3D spirals}
\begin{align}
    \begin{split}
        & x_t = \operatorname{sin}(\theta)\operatorname{cos}(\psi)\\
        & y_t = \operatorname{sin}(\theta)\operatorname{sin}(\psi)\\
        & z_t = \operatorname{cos}(\theta)\\
        & \dot{\theta}_t = 0.3(\pi - \theta_t)\\
        & \dot{\psi}_t = 0.3(2c\pi - \psi_t)\\
        & \psi_{t+1} = \psi_{t} + 0.003\dot{\psi}_t\\
        & \theta_{t+1} = \theta_{t} + 0.003\dot{\theta}_t
    \end{split}
    \label{eq:spiral3Dstable}
\end{align}
For each $c$, trajectories were generated until the distance between subsequent points fell below $1e-2$.
 
\textbf{2-dimensional Archimedean Spiral}
\begin{align}
    \begin{split}
        & \theta_{\text{goal}} = 3\pi\\
        & R = 0.1m \text{  (initial radius of spiral)}\\
        & \dot{\theta} = \frac{\theta_{\text{goal}}}{T}\\
        & x_t = r_t\operatorname{cos}(\theta_t)\\
        & y_t = r_t\operatorname{sin}(\theta_t)\\
        & \theta_{t+1} = \dot{\theta}(t+1) \\
        & r_{t+1} = R\frac{\theta_{t+1}}{\theta_{\text{goal}}}
    \end{split}
    \label{eq:archimideanSpiral}
\end{align}
for $t \in [0, 12]$.

\end{appendices}
 
\printbibliography

@article{schaal_is_1999,
	title = {Is imitation learning the route to humanoid robots?},
	volume = {3},
	issn = {13646613},
	url = {https://linkinghub.elsevier.com/retrieve/pii/S1364661399013273},
	doi = {10.1016/S1364-6613(99)01327-3},
	language = {en},
	number = {6},
	urldate = {2021-08-13},
	journal = {Trends in Cognitive Sciences},
	author = {Schaal, Stefan},
	month = jun,
	year = {1999},
	pages = {233--242},
}

@article{ravichandar_recent_2020-1,
	title = {Recent {Advances} in {Robot} {Learning} from {Demonstration}},
	volume = {3},
	issn = {2573-5144, 2573-5144},
	url = {https://www.annualreviews.org/doi/10.1146/annurev-control-100819-063206},
	doi = {10.1146/annurev-control-100819-063206},
	language = {en},
	number = {1},
	urldate = {2021-08-13},
	journal = {Annual Review of Control, Robotics, and Autonomous Systems},
	author = {Ravichandar, Harish and Polydoros, Athanasios S. and Chernova, Sonia and Billard, Aude},
	month = may,
	year = {2020},
	pages = {297--330},
}

@article{mohammad_khansari-zadeh_learning_2014,
	title = {Learning control {Lyapunov} function to ensure stability of dynamical system-based robot reaching motions},
	volume = {62},
	issn = {09218890},
	url = {https://linkinghub.elsevier.com/retrieve/pii/S0921889014000372},
	doi = {10.1016/j.robot.2014.03.001},
	language = {en},
	number = {6},
	urldate = {2021-07-19},
	journal = {Robotics and Autonomous Systems},
	author = {Mohammad Khansari-Zadeh, S. and Billard, Aude},
	month = jun,
	year = {2014},
	pages = {752--765},
}

@article{khansari-zadeh_learning_2011,
	title = {Learning {Stable} {Nonlinear} {Dynamical} {Systems} {With} {Gaussian} {Mixture} {Models}},
	volume = {27},
	issn = {1552-3098, 1941-0468},
	url = {http://ieeexplore.ieee.org/document/5953529/},
	doi = {10.1109/TRO.2011.2159412},
	language = {en},
	number = {5},
	urldate = {2021-07-19},
	journal = {IEEE Transactions on Robotics},
	author = {Khansari-Zadeh, S. Mohammad and Billard, Aude},
	month = oct,
	year = {2011},
	pages = {943--957},
}

@article{kronander_passive_2016,
	title = {Passive {Interaction} {Control} {With} {Dynamical} {Systems}},
	volume = {1},
	issn = {2377-3766, 2377-3774},
	url = {http://ieeexplore.ieee.org/document/7358081/},
	doi = {10.1109/LRA.2015.2509025},
	language = {en},
	number = {1},
	urldate = {2021-06-22},
	journal = {IEEE Robotics and Automation Letters},
	author = {Kronander, Klas and Billard, Aude},
	month = jan,
	year = {2016},
	pages = {106--113},
}

@article{perrin_fast_2016,
	title = {Fast diffeomorphic matching to learn globally asymptotically stable nonlinear dynamical systems},
	volume = {96},
	issn = {01676911},
	url = {https://linkinghub.elsevier.com/retrieve/pii/S0167691116300846},
	doi = {10.1016/j.sysconle.2016.06.018},
	language = {en},
	urldate = {2021-06-22},
	journal = {Systems \& Control Letters},
	author = {Perrin, Nicolas and Schlehuber-Caissier, Philipp},
	month = oct,
	year = {2016},
	pages = {51--59},
}

@inproceedings{rana_euclideanizing_2020,
  title={Euclideanizing flows: Diffeomorphic reduction for learning stable dynamical systems},
  author={Rana, Muhammad Asif and Li, Anqi and Fox, Dieter and Boots, Byron and Ramos, Fabio and Ratliff, Nathan},
  booktitle={Learning for Dynamics and Control},
  pages={630--639},
  year={2020},
  organization={PMLR}
}

@article{khansari-zadeh_learning_2011-1,
	title = {Learning {Stable} {Nonlinear} {Dynamical} {Systems} {With} {Gaussian} {Mixture} {Models}},
	volume = {27},
	issn = {1552-3098, 1941-0468},
	url = {http://ieeexplore.ieee.org/document/5953529/},
	doi = {10.1109/TRO.2011.2159412},
	language = {en},
	number = {5},
	urldate = {2021-06-22},
	journal = {IEEE Transactions on Robotics},
	author = {Khansari-Zadeh, S. Mohammad and Billard, Aude},
	month = oct,
	year = {2011},
	pages = {943--957},
}

@article{neumann_learning_2015,
	title = {Learning robot motions with stable dynamical systems under diffeomorphic transformations},
	volume = {70},
	issn = {09218890},
	url = {https://linkinghub.elsevier.com/retrieve/pii/S0921889015000883},
	doi = {10.1016/j.robot.2015.04.006},
	language = {en},
	urldate = {2021-06-22},
	journal = {Robotics and Autonomous Systems},
	author = {Neumann, Klaus and Steil, Jochen J.},
	month = aug,
	year = {2015},
	pages = {1--15},
}

@inproceedings{rai2014learning,
  title={Learning coupling terms for obstacle avoidance},
  author={Rai, Akshara and Meier, Franziska and Ijspeert, Auke and Schaal, Stefan},
  booktitle={2014 IEEE-RAS International Conference on Humanoid Robots},
  pages={512--518},
  year={2014},
  organization={IEEE}
}

@article{joshi2000landmark,
  title={Landmark matching via large deformation diffeomorphisms},
  author={Joshi, Sarang C and Miller, Michael I},
  journal={IEEE transactions on image processing},
  volume={9},
  number={8},
  pages={1357--1370},
  year={2000},
  publisher={IEEE}
}

@article{chaandar2019learning,
  title={Learning position and orientation dynamics from demonstrations via contraction analysis},
  author={chaandar Ravichandar, Harish and Dani, Ashwin},
  journal={Autonomous Robots},
  volume={43},
  number={4},
  pages={897--912},
  year={2019},
  publisher={Springer}
}

@inproceedings{Salvador2004FastDTWTA,
  title={FastDTW: Toward Accurate Dynamic Time Warping in Linear Time and Space},
  author={Stan Salvador and Philip Ka-Fai Chan},
  year={2004}
}

@article{tee2007eigenvectors,
  title={Eigenvectors of block circulant and alternating circulant matrices},
  author={Tee, G. J.},
  journal={New Zealand Journal of Mathematics},
  volume={36},
  number={8},
  pages={195--211},
  year={2007}
}

@book{bapat2010graphs,
  title={Graphs and matrices},
  author={Bapat, Ravindra B.},
  volume={27},
  year={2010},
  publisher={Springer}
}

@book{horn2012matrix,
  title={Matrix analysis},
  author={Horn, R. A. and Johnson, C. R.},
  year={2012},
  publisher={Cambridge university press}
}

@article{belkin2003laplacian,
  title={Laplacian eigenmaps for dimensionality reduction and data representation},
  author={Belkin Mikhail and Niyogi Partha},
  journal={Neural computation},
  volume={15},
  number={6},
  pages={1373--1396},
  year={2003},
  publisher={MIT Press}
}

@inproceedings{figueroa2018physically,
  title={A Physically-Consistent Bayesian Non-Parametric Mixture Model for Dynamical System Learning.},
  author={Figueroa, Nadia and Billard, Aude},
  booktitle={CoRL},
  pages={927--946},
  year={2018}
}

@inproceedings{NEURIPS2019_0a4bbced,
 author = {Kolter, J. Zico and Manek, Gaurav},
 booktitle = {Advances in Neural Information Processing Systems},
 editor = {H. Wallach and H. Larochelle and A. Beygelzimer and F. d\textquotesingle Alch\'{e}-Buc and E. Fox and R. Garnett},
 pages = {},
 publisher = {Curran Associates, Inc.},
 title = {Learning Stable Deep Dynamics Models},
 url = {https://proceedings.neurips.cc/paper/2019/file/0a4bbceda17a6253386bc9eb45240e25-Paper.pdf},
 volume = {32},
 year = {2019}
}

@inproceedings{urain2020imitationflow,
  title={ImitationFlow: Learning Deep Stable Stochastic Dynamic Systems by Normalizing Flows},
  author={Urain, Julen and Ginesi, Michele and Tateo, Davide and Peters, Jan},
  booktitle={2020 IEEE/RSJ International Conference on Intelligent Robots and Systems (IROS)},
  pages={5231--5237},
  year={2020},
  organization={IEEE}
}

@inproceedings{ravichandar2017learning,
  title={Learning partially contracting dynamical systems from demonstrations},
  author={Ravichandar, Harish and Salehi, Iman and Dani, Ashwin},
  booktitle={Conference on Robot Learning},
  pages={369--378},
  year={2017},
  organization={PMLR}
}

@misc{FicheraBillard2022,
  doi = {10.48550/ARXIV.2202.09171},
  url = {https://arxiv.org/abs/2202.09171},
  author = {Fichera, Bernardo and Billard, Aude},
  title = {Linearization and Identification of Multiple-Attractors Dynamical System through Laplacian Eigenmaps},
  publisher = {arXiv},
  year = {2022},
  copyright = {Creative Commons Attribution 4.0 International}
}

\end{document}